\documentclass[10pt,twocolumn,letterpaper]{article}

\usepackage[pagenumbers]{cvpr}              

\usepackage{graphicx}
\usepackage{amsmath}
\usepackage{amssymb}
\usepackage{booktabs}
\usepackage[font=small]{subcaption}
\usepackage[ruled,vlined]{algorithm2e}
\usepackage{adjustbox}
\usepackage{lscape}

\usepackage{array,xcolor,colortbl}
\usepackage{multirow}
\newcolumntype{C}{>{\centering\arraybackslash}p{0.25\textwidth}}

\renewcommand{\paragraph}[1]{\vspace{3pt}\noindent\textbf{#1}~}

\usepackage[pagebackref,breaklinks,colorlinks]{hyperref}

\usepackage[capitalize]{cleveref}
\crefname{section}{Sec.}{Secs.}
\Crefname{section}{Section}{Sections}
\Crefname{table}{Table}{Tables}
\crefname{table}{Tab.}{Tabs.}

\definecolor{rred}{rgb}{1,0,0}
\definecolor{red}{rgb}{0.95,0.4,0.4}
\definecolor{blue}{rgb}{0.4,0.4,0.95}
\definecolor{darkgreen}{rgb}{0.15,0.6,0.15}
\definecolor{grey}{rgb}{0.6,0.6,0.6}

\def\cvprPaperID{8877}
\def\confName{CVPR}
\def\confYear{2022}

\newcommand{\algoHRule}[0]{\vspace{3pt}\hrule\vspace{3pt}}

\begin{document}

\title{Learning Structured Gaussians to Approximate Deep Ensembles}

\author{Ivor J.A. Simpson\thanks{Accepted at CVPR 2022}\\
University of Sussex, UK\\
{\tt\small i.simpson@sussex.ac.uk}
\and
Sara Vicente\\
Niantic, UK\\
{\tt\small svicente@nianticlabs.com}

\and
Neill D.F. Campbell \\
University of Bath, UK \\
{\tt\small n.campbell@bath.ac.uk}
}
\maketitle

\newcommand{\mLabel}[0]{{d}}
\newcommand{\mLogit}[0]{{d}}
\newcommand{\mInput}[0]{{x}}
\newcommand{\mLabels}[0]{\mathbf{d}}
\newcommand{\mLogits}[0]{\mathbf{d}}
\newcommand{\mInputs}[0]{\mathbf{x}}
\newcommand{\mModelParams}[0]{\mathbf{w}}
\newcommand{\mDataset}[0]{\mathcal{D}}

\newcommand{\mAllLogitsSamples}[0]{\mathbf{\tilde D}}

\newcommand{\mMean}[0]{\boldsymbol{\mu}}
\newcommand{\mPrecision}[0]{\boldsymbol{\Lambda}}
\newcommand{\mCovariance}[0]{\boldsymbol{\Sigma}}
\newcommand{\mCholPrec}[0]{\mathbf{L}_{\Lambda}}
\newcommand{\mCholCovar}[0]{\mathbf{L}_{\Sigma}}

\newcommand{\mNumImages}[0]{I}
\newcommand{\mNumLabels}[0]{K}
\newcommand{\indLabel}[0]{k}
\newcommand{\mNumPixels}[0]{N}
\newcommand{\indPix}[0]{n}
\newcommand{\mNumSamples}[0]{S}
\newcommand{\indSamp}[0]{s}
\newcommand{\mNumJacobiIter}[0]{J}
\newcommand{\indJacobi}[0]{j}

\newcommand{\given}[0]{\,\vert\,}
\newcommand{\inv}[1]{#1^{-1}}
\newcommand{\trans}[1]{#1^{\top}}
\newcommand{\transInv}[1]{#1^{-\top}}
\newcommand{\mD}[1]{\,\mathrm{d}#1\,}
\newcommand{\mNormal}[0]{\mathcal{N}}
\newcommand{\mExpect}[0]{\mathbb{E}}
\newcommand{\mSoftmax}[0]{\textsc{Softmax}}
\newcommand{\mCat}[0]{\textsc{Cat}}
\newcommand{\mDiag}[0]{\textrm{diag}}
\newcommand{\bigO}[0]{\mathcal{O}}

\begin{abstract}
This paper proposes using a sparse-structured multivariate Gaussian to provide a closed-form approximator for the output of probabilistic ensemble 
 models used for dense image prediction tasks. 
This is achieved through a convolutional neural network that predicts the mean and covariance of the distribution, where the inverse covariance is parameterised by a sparsely structured Cholesky matrix. 
Similarly to distillation approaches, our single network is trained to maximise the probability of samples from pre-trained probabilistic models, in this work we use a fixed ensemble of networks.
Once trained, our compact representation can be used to efficiently draw spatially correlated samples from the approximated output distribution.
Importantly, this approach captures the uncertainty and structured correlations in the predictions explicitly in a formal distribution, rather than implicitly through sampling alone. 
This allows direct introspection of the model, enabling visualisation of the learned structure.
Moreover, this formulation provides two further benefits: estimation of a sample probability, and the introduction of arbitrary spatial conditioning at test time.
We demonstrate the merits of our approach on monocular depth estimation and show that the advantages of our approach are obtained with comparable quantitative performance.
\end{abstract}


\section{Introduction}\label{sec:introduction}

\begin{figure}[t]
    \centering
    \includegraphics[width=0.45\textwidth]{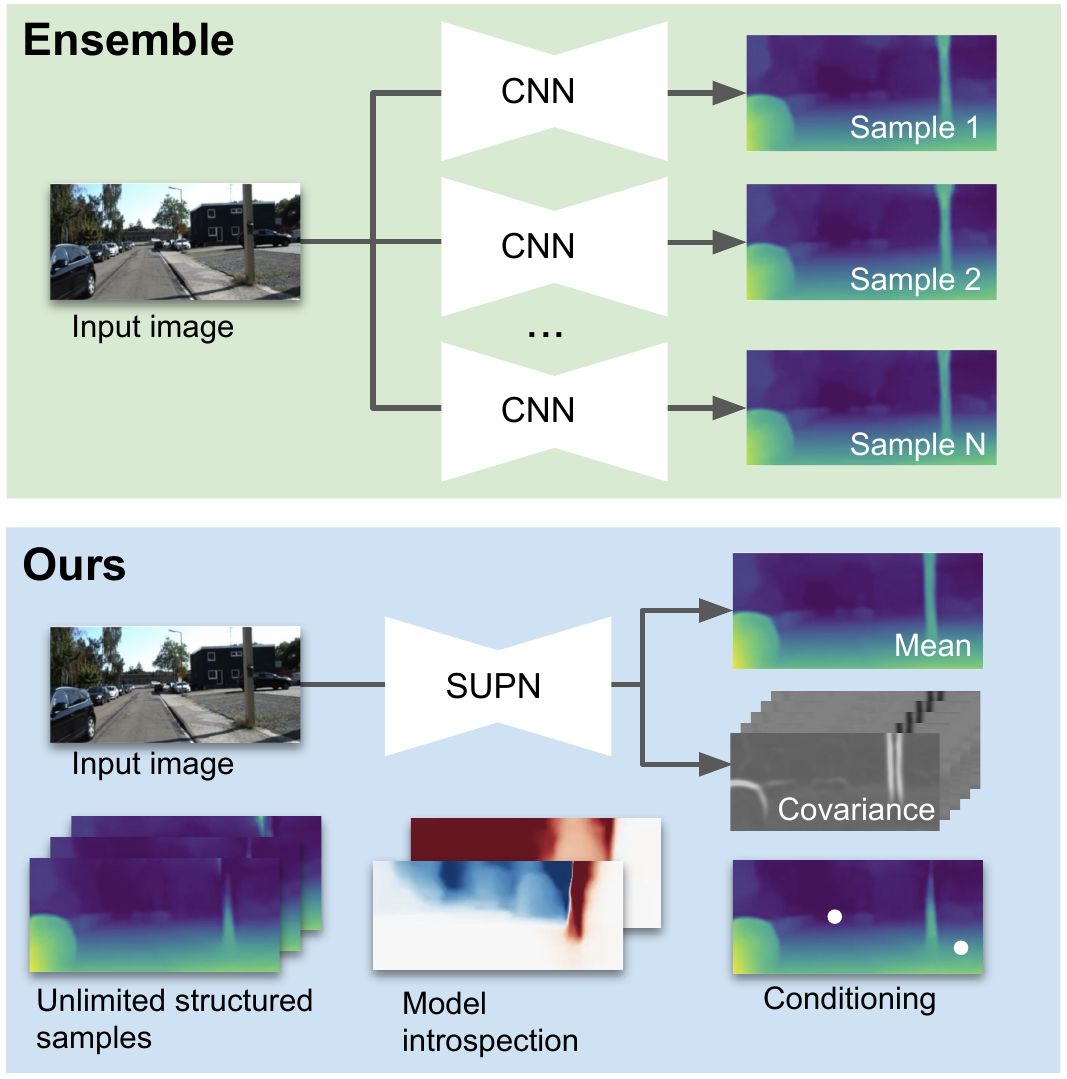}
    \caption{Our method is trained to approximate the output of an ensemble, by using structured uncertainty prediction networks (SUPN) to predict a mean and covariance for a multivariate Gaussian distribution. This explicit distribution enables a variety of tasks including: sampling, conditioning and model introspection. }
    \label{fig:teaser}
\end{figure}

Single prediction neural networks are ubiquitous in computer vision and have demonstrated extensive capability for a variety of tasks.
However, researchers are increasingly interested in capturing the uncertainty in estimation tasks to combat over-confidence and ambiguity; such concerns are important when building robust systems that connect computer vision approaches to down-stream applications.
The deployment of neural networks for safety-critical tasks, such as autonomous driving, requires an accurate measure of uncertainty. 
While Bayesian Neural Networks~\cite{mackay1995probable} are often a model of choice for uncertainty estimation, ensembles~\cite{lakshminarayanan2017simple} have been proposed as a simple alternative.
Empirically, ensembles have been shown to produce good measures of uncertainty for vision tasks~\cite{li2020improving, poggi2020uncertainty} and allow practitioners to exploit associated application-specific inductive biases, for example established architectures, directly.

\paragraph{Limitations of implicit approaches}
Despite their popularity, ensembles have a number of drawbacks that we group into three themes.
Firstly, they come at an increased cost compared with deterministic networks. At training time, they require training multiple deep models, while at test time, multiple inference passes are required. MC-dropout \cite{gal2016dropout} saves computation at training time, but it still requires multiple passes at inference time. 
Secondly, these approaches only provide an implicit distribution over probable model outputs. Any uncertainty captured is only accessible through ancestral sampling.
Accordingly, one cannot calculate conditional samples, or assess the likelihood of a new sample given the learned model. 
Finally, and of increasing importance to the community, introspection of the trained models is very difficult. 

When combining computer vision with larger systems, there is virtue to summarising the posterior distribution in a formal and compact form that can be visualised and appropriately used to inform downstream tasks.
The computational challenges have prompted work on producing a single model to approximate the output of an ensemble; so called ``ensemble distillation''~\cite{malinin2020ensemble, li2020improving,bulo2016dropout,shen2021real}. 

\paragraph{Lack of structure in distillation methods}
Previous methods focus on: classification problems~\cite{malinin2020ensemble, li2020improving}, approximating only the mean of the ensemble~\cite{bulo2016dropout}, or modelling independent per-pixel variance~\cite{shen2021real}.
In contrast, while we also adopt a single model to reduce the computational cost, we propose to learn a model that approximates the ensemble by formally capturing structure in the output space; this is more appropriate for dense prediction tasks. 
When making per-pixel predictions, it is common to use models that capture spatial correlation in the output space. In particular, models such as Markov or Conditional Random Fields~\cite{prince2012computer}, which capture correlations between neighbouring pixels, have been extensively used in computer vision.
However, capturing the structure of the output space is less explored in the context of modelling uncertainty.
Previous work has focused on per-pixel heteroscedastic uncertainty, by using a Gaussian~\cite{kendall2017uncertainties, shen2021real} or Laplace~\cite{klodt2018supervising} likelihood model with diagonal covariance.
Since these models do not capture correlations between pixels, samples suffer from salt-and-pepper (independent) noise.

\paragraph{Capturing structure explicitly}
Previously, adopting per-pixel uncertainty representations usually follows from the expectation that direct estimation of a full covariance structure is intractable in both storage $\mathcal{O}(\text{pixels}^2)$ and computation $\mathcal{O}(\text{pixels}^3)$.
Recently, however, Dorta~\etal~\cite{dorta2018structured} introduced Structured Uncertainty 
Pprediction Networks (SUPN) for generative models. The paper extended a Variational Auto Encoder (VAE)~\cite{kingma2013auto} with a likelihood model that is Gaussian with a full covariance matrix. 
The authors show how this can be predicted efficiently by using a sparse approximation of the Cholesky decomposition of the \emph{precision} matrix. 
Working in the domain of the precision allows a dense covariance structure to be obtained whilst also respecting our prior that long range structure is derived from the propagation of local image statistics.
By encoding a full covariance matrix, the samples obtained from such a model capture these long range correlations in the image domain and are free from salt-and-pepper (independent) noise.

\paragraph{Contributions} 
In this work, we build on SUPN~\cite{dorta2018structured} and show that a deep network can be trained in a regression setting to predict a structured Gaussian distribution that approximates the output distribution of methods that capture model uncertainty, such as ensembles~\cite{lakshminarayanan2017simple} and MC-dropout~\cite{gal2016dropout}.
We introduce a novel efficient approach to drawing conditioned or unconditioned samples from a structured multivariate Gaussian distribution with a sparsely structured precision matrix.
By taking full advantage of the closed form nature of the Gaussian distribution, our method allows introspection and enables conditioning at test time, which proves cumbersome for other methods.
Importantly, our approach is not limited to Gaussian likelihoods over the prediction space (see~\S~\ref{sec:non-gauss-likelihood}).

\paragraph{Evaluation}
We demonstrate the efficacy of our method for the task of depth estimation. 
Experiments show that the new advantages can be obtained without sacrificing quantitative performance, with results comparable to the original ensemble; we consider metrics over both accuracy and the capture of uncertainty.
The samples are found to follow the ensembles without being limited in the number that can be drawn. 
The compact representation is capable of encoding a rich distribution with only a modest increase in computation over a single deterministic network.
Furthermore, we demonstrate using our explicit representation to perform conditional sampling and illustrate the ability to inspect the model and visualise the correlation structure learned.

\section{Background}

Our goal is to model $p(\mLabels \given \mInputs)$, where $\mInputs$ is the observed image and $\mLabels$ is a per-pixel prediction, \eg a semantic labelling or depth map. While most deterministic deep models can be seen as capturing the mean $\mMean(\mInputs)$ of this distribution, we are interested in models that also capture the variance $\mCovariance(\mInputs)$.

\subsection{Uncertainty in Deep Models}

Previous work for probabilistic modelling using neural networks, can be broadly grouped into three categories: (1) Bayesian approaches that model uncertainty of the network parameters, (2) methods that empirically approximate Bayesian approaches by predicting multiple hypothesis, and (3) approaches that model $p(\mLabels \given \mInputs)$ directly by predicting a parametric distribution.
The literature on uncertainty modelling in neural networks is vast and we direct the interested reader to a recent review~\cite{abdar2020review}.

\paragraph{Modelling uncertainty in the parameters}
Bayesian neural networks~\cite{mackay1995probable} model uncertainty by modelling the probability distribution of the learned weights~$\mModelParams$ of the network. 
The resulting posterior $p(\mLabels \given \mInputs)$ is then obtained by marginalising over the weights:
\begin{equation}
    p(\mLabels \given \mInputs, \mDataset) = \int p(\mLabels \given \mInputs, \mModelParams) \, p(\mModelParams \given \mDataset)~d \mModelParams \,,
\label{eq:bayesian_networks}
\end{equation}
where $\mModelParams$ are the model parameters, and we make explicit the dependence on the dataset $\mDataset$.

While this approach is able to model arbitrary distributions $p(\mLabels|\mInputs)$,
and generate samples which are correlated in output space, it also suffers some limitations. 
The majority of approaches rely on mean-field approximations over the weights to maintain tractability. 
In addition, it is difficult to condition on any of the output values due to the absence of a parametric distribution over the posterior. %

MC-dropout~\cite{gal2016dropout} approximates Bayesian networks by using dropout at both training and test time. Dropout was first proposed to reduce over-fitting in deep neural networks~\cite{srivastava2014dropout} and it proceeds by randomly setting some of the weights of the network to zero. It has been shown, \cite{gal2016dropout}, that this random dropping of weights at test time is akin to sampling from the distribution $p(\mModelParams \given \mDataset)$ and may be used to approximate the integral in~ \eqref{eq:bayesian_networks}.

\paragraph{Multiple hypothesis}
Ensemble methods make use of multiple models and combine them to get a single prediction. 
Deep ensembles can be trained using bootstraping~\cite{lakshminarayanan2017simple}, \ie splitting the training set into multiple random subsets and training each model in the ensemble independently. Alternatively, to save computation, a deep ensemble can be trained by taking multiple snapshots~\cite{huang2017snapshot} from the same training procedure, requiring a cyclic learning rate.
Ensembles have been shown to provide good measures of uncertainty~\cite{lakshminarayanan2017simple}. They can be seen as approximating Bayesian networks by replacing the integral in~\eqref{eq:bayesian_networks} by a sum over a discrete number of models.
As discussed in \S~\ref{sec:introduction}, training and inference procedures can become expensive in terms of maintaining an increasing number of networks; practical approaches are often limited in the number of distinct models that, in turn, restricts the number test-time samples.

\paragraph{Predictive uncertainty via parametric distributions}
The other alternative to modelling uncertainty is to use a feed-forward neural network to predict the parameters of a parametric distribution~\cite{kendall2017uncertainties}.
For regression tasks, $p(\mLabels|\mInputs)$ is typically described by a Gaussian likelihood, where the mean and variance are outputs of the neural network:
\begin{equation}
    p(\mLabels \given \mInputs) \sim  \mNormal\big(\mLabels \given \mMean(\mInputs), \mCovariance(\mInputs) \big) \,,
\label{eq:gaussianl_likelihood}
\end{equation}
where $\mCovariance(\mInputs)$ is usually approximated by a diagonal matrix where the diagonal elements are predicted by the network.
Kendall and Gal~\cite{kendall2017uncertainties} discuss how the predicted variance can be seen as a loss attenuation factor, reducing the loss for outliers; this predicted per-pixel variance is shown to correlate with error in the predictions.
Evaluating predictive uncertainty is more efficient since a single pass of the network at test time is sufficient to fully determine the measure of uncertainty.
However, independent per-pixel uncertainty estimates fail to capture spatial correlation that is known to exist in images; samples from these models are destined to be unrealistic and suffer from salt-and-pepper noise.

\paragraph{Ensemble distillation}
Recently, there has been a growing interest in approximating the probabilistic output of an ensemble by a single model\cite{malinin2020ensemble, li2020improving,bulo2016dropout,shen2021real}. This process is commonly named ``distillation''. Most of the focus has been on classification \cite{malinin2020ensemble, li2020improving}, where the goal is to predict the class of the image. While these methods show impressive results in detecting out-of-distribution images, they are not easily extended for dense prediction tasks.
Other methods focus on approximating only the mean of the ensemble distribution \cite{bulo2016dropout}, or modelling independent per-pixel variance \cite{shen2021real}.
In contrast, our model also does ensemble distillation, but can capture structure in the output space.

\paragraph{Uncertainty in depth prediction models}
The goal of self-supervised depth estimation is to train a network to predict single image depth maps without explicit depth supervision \cite{godard2017monodepth, godard2019digging}. Instead, self-supervised approaches use geometric constraints between two calibrated stereo cameras to learn depth prediction. At test time, these methods do not require a stereo pair, only a single image.
Given the inherit ambiguity of predicting depth from a single image, depth prediction is a natural use case for uncertainty estimation in dense prediction tasks.
In \cite{poggi2020uncertainty} the authors review and compare different approaches for uncertainty prediction for self-supervised depth prediction. They focus on methods that predict multiple hypothesis, such as dropout~\cite{gal2016dropout} and ensembles~\cite{lakshminarayanan2017simple}, methods that predict per-pixel independent heteroscedastic uncertainty~\cite{klodt2018supervising}, and combinations of both.

In the experiments, we use the pre-trained networks provided by \cite{poggi2020uncertainty} to evaluate the efficiency of our method in approximating ensembles.
In particular, we use their most successful model, which combines an ensemble with predictive uncertainty. Their ensembles are trained using bootstrapping \cite{lakshminarayanan2017simple}, and they use an uncorrelated Laplace distribution for predictive parametric uncertainty.

Xia~\etal~\cite{xia2020generating} show how a probabilistic model for depth prediction can be explored by downstream tasks such as inference with additional information. They model uncertainty at a patch level in a model akin to a Markov Random Field.
In contrast to our approach, the method requires solving a complex optimization problem at inference time.

\subsection{Predicting Structured Gaussian Distributions}
\label{sec:SUPN}
To approximate an ensemble, we train a network to predict the parameters of a Gaussian distribution.
Given an input image $\mInputs$ the network outputs the parameters of a Gaussian distribution $\mMean(\mInputs)$ and $\mCovariance(\mInputs)$.
We focus on dense prediction tasks. For these tasks, if $N$ is the number of pixels in the input image, the size of $\mMean$ is also $N$ while a full $\mCovariance$ matrix has $N^{2}$ parameters.
The quadratic scaling of the number of parameters of the covariance matrix leads to the common remedy of a diagonal matrix, which requires only $N$ parameters. However, this simplifying assumption prohibits the capture of correlations between pixels.

\paragraph{Structured Uncertainty Prediction Networks} %
Our approach builds on the work, \cite{williams1996using, dorta2018structured}, where the parameterisation used is the Cholesky decomposition of the precision matrix, \ie the network predicts $\mCholPrec$ directly, where $\mCholPrec \trans{\mCholPrec} = \inv{\mCovariance}$ and $\mCholPrec$ is a lower triangular matrix. 
For completeness, we review some of the properties of the parameterisation presented in \cite{dorta2018structured}, which we use in our work.

When choosing a parameterisation, there are a few criteria that should be taken into account: how easy it is to evaluate the likelihood function required for training, how easy it is to sample from the distribution at inference time and how easy is to impose that the covariance matrix (or equivalently precision matrix) is symmetric and positive definite?
Direct prediction of the Cholesky factor guarantees that the precision matrix is symmetric. 
To guarantee that it is positive definite, the diagonal values of the Cholesky decomposition are required to be positive; an easy constraint to enforce in practice. 
This choice of parameterisation allows for easy computation of the log-likelihood of the multivariate Gaussian distribution. 
However, sampling is more difficult to perform, since access to the covariance is required. 
We discuss a new efficient method for sampling in \S~\ref{sec:sampling}.

\paragraph{Sparsity} Despite the advantages of using this parameterisation and the fact that the Cholesky is a lower triangular matrix, the number of elements still grows quadratically with respect to the number of pixels, $N$, making it prohibitive to directly estimate for large images.
We follow SUPN~\cite{dorta2018structured} in imposing sparsity in the Cholesky matrix $\mCholPrec$.
For each pixel, we only populate the Cholesky matrix for pixels which are in a small neighborhood, while keeping the matrix lower-triangular. We include an illustration in the supplemental material.
This sparse Cholesky matrix can be compactly represented by only predicting the non-zero values; for a $3 \times 3$ neighborhood, this corresponds to predicting the diagonal map plus 4 off-diagonal maps. 

Importantly, 
this representation can be encoded efficiently into popular APIs such as Tensorflow and PyTorch using standard convolutional operations.

\paragraph{Deep Gaussian MRFs} 
Our model can be seen as a Gaussian Markov Random Field, since the sparsity pattern on the precision matrix directly implies the Markov property: a variable is conditional independent of all other variables given its neighbours.
Similar to our approach, \cite{chandra2016fast, vemulapalli2016gaussian, rtf2012} use a regression model to predict the parameters of a Gaussian Conditional Random Field that captures structure in output space. They show improved results for semantic segmentation. However, they focus on predicting the MAP solution and %
do not make use of the full probability distribution.

\newcommand{\mIdentity}[0]{\mathbf{I}}

\section{Method}
Our goal is to train a single network that approximates the multiple outputs of an ensemble. We assume this ensemble is given as a pre-trained network(s), %
\eg from \cite{lakshminarayanan2017simple} or \cite{huang2017snapshot}.
We predict a structured multivariate Gaussian using the sparse representation discussed in \S~\ref{sec:SUPN}.

\subsection{Training}
Given %
$I$ training images $\{\mInputs_i \given i \in [1, \mNumImages]\}$,
the pre-trained ensemble is run for the full training set, to obtain $S$ distinct predictions per image $\{\mLabels_i^s \given s \in [1, S] \}$, where $S$ is the size of the ensemble or number of MC-dropout samples.

\paragraph{Log-likelihood loss}
Our network is trained to %
minimise the negative log-likelihood of the training set:
\begin{equation}
    \mathcal{L} = - \sum_{i=1}^{\mNumImages}\sum_{s=1}^{S} \, \log \mNormal\big(\mLabels_i^s \given \mMean(\mInputs_i), \mCovariance(\mInputs_i) \big) \,,
\end{equation}
where  $\mNormal\big(\mLabels_i^s \given \mMean(\mInputs_i), \mCovariance(\mInputs_i) \big)$ is the probability density function of a multivariate Gaussian distribution.

\subsection{Inference}
In common with ensembles and MC-dropout, we can use our model to obtain samples from the predictive distribution $p(\mLabels \given \mInputs)$. In contrast with ensembles, our model is not restricted on the number of samples that can be taken; we discuss an efficient sampling procedure in~\S~\ref{sec:sampling}.
More importantly, since our model predicts a closed form probability function, it allows for additional inference tasks which are not possible with ensembles or MC-dropout.

\paragraph{Evaluation of the predictive log-likelihood}
Our model allows evaluating the log-likelihood for a given dense prediction. This is useful for model comparison.

\paragraph{Conditional distribution}
The output Gaussian distribution can be used to compute the conditional distribution of some pixel labels, given the label for other pixels.
The ability of drawing conditional samples has practical applications, for example: for depth completion, where the depth of a small number of pixels is provided by an external sensor, such as a LIDAR scanner; or for interactive image segmentation, where the label of a few pixels is provided by a user.

\subsection{Efficient Sampling}\label{sec:sampling}
Sampling from a Multivariate Gaussian distribution with a diagonal covariance matrix $\mCovariance = \mDiag(\sigma_1, \dots, \sigma_{\mNumPixels})$ can proceed with a straight forward sampling approach where each dimension (pixel) is independent:
\newcommand{\mLogitSample}[0]{\tilde\mLogit^{(\indSamp)}}%
\begin{equation}
    \mLogitSample_{\indPix} = \mMean_{\indPix} + \sigma_{\indPix} \, \tilde\varepsilon^{(\indSamp)}_{\indPix} , \quad \tilde\varepsilon^{(\indSamp)}_{\indPix} \sim \mNormal(0, 1) \, .
\end{equation}

If the Gaussian distribution has a general covariance, however, then the sample cannot be computed independently for each pixel and must be drawn through a square root matrix of the covariance, such as the Cholesky factor:
\newcommand{\mLogitsSample}[1]{\tilde\mLogits^{(#1)}}%
\newcommand{\mNoiseSample}[1]{\boldsymbol{\tilde\varepsilon}^{(#1)}}
\begin{equation}
    \mLogitsSample{\indSamp} = \mMean + \mCholCovar \, \mNoiseSample{\indSamp} , \quad \mNoiseSample{\indSamp} \sim \mNormal(\mathbf{0}, \mIdentity_{\mNumPixels}) \, ,
\label{eq:loss}
\end{equation}
where $\mCholCovar \trans{\mCholCovar} = \mCovariance$. 
Computation of the dense covariance matrix from the sparse precision, followed by the Cholesky operation would involve a computational complexity of $\bigO(\mNumPixels^3)$ and $\bigO(\mNumPixels^2)$ storage making it infeasible.

\paragraph{Efficient calculation via the Jacobi method}
Fortunately, adopting a sparse structure over the Cholesky precision matrix $\mCholPrec$ means that we can perform a matrix multiplication efficiently.
We can exploit this to take approximate samples using a truncated (to $\mNumJacobiIter$ iterations) version of the Jacobi iterative solver to invert $\mCholPrec$. 
This results in a tractable algorithm for obtaining approximate samples of sufficient quality. We can take multiple samples from the same distribution simultaneously while retaining efficiency.

We start with a set of $\mNumSamples$ standard Gaussian samples,
\newcommand{\mRawSamples}[0]{\boldsymbol{\tilde E}}%
\newcommand{\mJacobiSamples}[0]{}%
\begin{align}
    \mRawSamples = [\mNoiseSample{1}, \dots, \mNoiseSample{\mNumSamples} ] , \quad \mNoiseSample{\indSamp} \sim \mNormal(\mathbf{0}, \mathbf{I}_{\mNumPixels}) \, .
\end{align}
We then note that the transposed, inverse of the precision Cholesky matrix can be used as a sampling matrix since
\begin{equation}
    \mCovariance = \inv{\mPrecision} = \inv{(\mCholPrec \trans{\mCholPrec})} = \transInv{\mCholPrec} \inv{\mCholPrec} \, ,
\end{equation}
indicating that $\transInv{\mCholPrec}$ is the LHS of a square root matrix for $\mCovariance$.
Thus we draw low variance Monte Carlo samples as
\begin{equation}
    \mAllLogitsSamples = [\mLogitsSample{1}, \dots, \mLogitsSample{\mNumSamples}] = \mMean + \transInv{\mCholPrec} \mRawSamples \, .
\end{equation}
\newcommand{\mAllZeroMeanJacobiSamples}[0]{\mathbf{S}}%
To invert $\trans{\mCholPrec}$ efficiently, we use $\mNumJacobiIter$ Jacobi iterations; these are particularly efficient to apply with a sparse matrix that is already lower triangular. 
We initialise $\mAllZeroMeanJacobiSamples^{(0)} = \mRawSamples$ and then, at each iteration, update the samples with
\newcommand{\mJacobiD}[0]{D_{\Lambda}}%
\newcommand{\mJacobiU}[0]{U_{\Lambda}}%
\begin{align}
    \mAllZeroMeanJacobiSamples^{(\indJacobi + 1)} \leftarrow \inv{\mJacobiD} \big( \mRawSamples - \mJacobiU \, \mAllZeroMeanJacobiSamples^{(\indJacobi)} \big) \, , \label{eqn:sampling_with_precision}
\end{align}
where $\mJacobiD := \mDiag(\trans{\mCholPrec})$ and $\mJacobiU := \trans{\mCholPrec} - \mJacobiD$, a strictly upper triangular matrix. The final samples are then obtained by the addition of the mean such that $\mAllLogitsSamples = \mMean + \mAllZeroMeanJacobiSamples$. 

\newcommand{\mFilterBank}[0]{\mathbf{f}}
\newcommand{\mAllFilterBanks}[0]{\mathbf{F}}
\newcommand{\mLogDiag}[0]{\boldsymbol{\phi}}
\newcommand{\mOffDiag}[0]{\boldsymbol{\psi}}
\newcommand{\mOffDiagShuffled}[0]{\boldsymbol{\psi}_{\mathrm{shuff}}}
\newcommand{\mNumFilters}[0]{L}

\begin{algorithm}[t]
\SetAlgoLined
\KwResult{Samples drawn from a correlated multivariate Gaussian (with sparse precision) 
}
 Samples: $\mAllZeroMeanJacobiSamples^{(0)} \leftarrow \mRawSamples \sim \mNormal(\mathbf{0}, \mIdentity_{N})$, $N := W \times H$\;
 Local connection filters: $\mAllFilterBanks = \{\mFilterBank_{l}\}_{l=1}^{\mNumFilters}$\;
 Log diagonal terms: $\mLogDiag \in \mathbb{R}^{N}$\;
 Off diagonal terms: $\mOffDiag \in \mathbb{R}^{\mNumFilters \times N}$\;
 
 \algoHRule
 
 \For{$j \leftarrow 0$ \KwTo $J-1$}{
    $\mathbf{V} \leftarrow \textrm{Conv2D}(\mAllZeroMeanJacobiSamples^{(j)}, \mAllFilterBanks)$\;
    $\mathbf{v} \leftarrow \sum_{l=1}^{L} [\mathbf{V} \odot \mOffDiag]_{n,l}$\;
    $\mAllZeroMeanJacobiSamples^{(j+1)} \leftarrow \big[\exp(\mLogDiag)\big]^{-1} \odot (\mRawSamples - \mathbf{v})$
 }
 
 \algoHRule
 
 Output: $\mAllZeroMeanJacobiSamples^{(J)} \approx \big( \transInv{\mCholPrec} \mRawSamples \big) \sim \mNormal(\mathbf{0}, \inv{\mPrecision})$\;
 
 \caption{Jacobi sampling for the multivariate Gaussian distribution}
 \label{algo:jacobi}
\end{algorithm}

\newcommand{\mLogitsKnown}[0]{\mLogits_{\mathrm{K}}}
\newcommand{\mLogitsUnknown}[0]{\mLogits_{\mathrm{U}}}
\newcommand{\mKnown}[0]{\boldsymbol{\alpha}}
\newcommand{\mUnknown}[0]{\boldsymbol{\beta}}

\newcommand{\mMask}[0]{\mathbf{m}}
\newcommand{\mMaskKnown}[0]{\mMask_{\mathrm{K}}}
\newcommand{\mMaskUnknown}[0]{\mMask_{\mathrm{U}}}

\paragraph{Efficient conditional sampling}
As we have a closed form representation of the output distribution:
\begin{equation}
    \mLogits \sim \mNormal(\mMean, \mCovariance), \;\mCovariance = \transInv{\mCholPrec} \inv{\mCholPrec} \, ,
\end{equation}
we can find the expression for a resulting conditional distribution where we specify values for a subset of the pixels and sample from the resulting distribution over the remaining pixels. 
Let us partition the pixels into a set of known values $\mLogitsKnown$ and unknown values $\mLogitsUnknown$; pixels (arbitrarily) belong to either one set or the other under a pixel-wise mask:
\begin{equation}
    [\mMaskKnown]_{n} = \left\{ \begin{array}{cl}
        1, & n \in \mathcal{K} \\
        0, & n \in \mathcal{U}
    \end{array} \right. ,\; \mMaskUnknown = 1 - \mMaskKnown \,.
\end{equation}
Thus, with slight abuse of notation, we recover the full set of values as $\mLogits = \mMaskKnown \odot \mLogitsKnown + \mMaskUnknown \odot \mLogitsUnknown$.
The conditional distribution for the unknown values, given that the known values $\mLogitsKnown = \mKnown$, is the Gaussian conditional density:
\begin{align}
    p(\mLogitsUnknown \given \mLogitsKnown = \mKnown) &\phantom{:}\sim \mNormal(\mathbf{b}, \mathbf{B}) \,,\\
    \mathbf{b} &:= \mMean_{\mathrm{U}} +  \mCovariance_{\mathrm{UK}} \mCovariance_{\mathrm{KK}}  (\mKnown - \mMean_{\mathrm{K}}) \,,\\ 
    \mathbf{B} &:= \mCovariance_{\mathrm{UU}} - \mCovariance_{\mathrm{UK}} \inv{\mCovariance_{\mathrm{KK}}} \mCovariance_{\mathrm{KU}} \, ,
\end{align}
where the subscripts dictate the appropriate partitions of the mean vector or blocks of the covariance matrix.

Evaluating this directly, in matrix form, would again be prohibitively expensive, especially considering the matrix inversions (from precision to covariance matrices). Thankfully we can use a modified form of the Jacobi sampling method combined with Matheron's rule for conditional sampling. Matheron's rule states that if $(\mathbf{a}, \mathbf{b})$ are samples from the joint distribution $p(\mLogitsKnown, \mLogitsUnknown)$ then the random variable $\mathbf{b}$ conditioned on $\mathbf{a} = \mKnown$ can be found by:
\begin{equation}
    (\mathbf{b} \mid \mathbf{a} = \mKnown) \leftarrow \mathbf{b} +  \mCovariance_{\mathrm{UK}} \inv{\mCovariance_{\mathrm{KK}}} (\mKnown - \mathbf{a}) \,. \label{eqn:matheron_rule}
\end{equation}
We can use straight forward identities to convert Matheron's rule into an update equation in terms of the precision:
\begin{align}
    \begin{bmatrix} 
        \mPrecision_{\mathrm{KK}} & \mPrecision_{\mathrm{KU}} \\
        \mPrecision_{\mathrm{UK}} & \mPrecision_{\mathrm{UU}} 
    \end{bmatrix} \cdot
    \begin{bmatrix} 
        \mCovariance_{\mathrm{KK}} & \mCovariance_{\mathrm{KU}} \\
        \mCovariance_{\mathrm{UK}} & \mCovariance_{\mathrm{UU}} 
    \end{bmatrix} =&
    \begin{bmatrix} 
        \mIdentity & \mathbf{0} \\
        \mathbf{0} & \mIdentity 
    \end{bmatrix} \,, \\
    \Rightarrow \mPrecision_{\mathrm{UK}} \, \mCovariance_{\mathrm{KK}} + \mPrecision_{\mathrm{UU}}  \, \mCovariance_{\mathrm{UK}} =& \, \mathbf{0} \\
    \Rightarrow \mCovariance_{\mathrm{UK}} \, \inv{\mCovariance_{\mathrm{KK}}} = - \inv{\mPrecision_{\mathrm{UU}}} \, \mPrecision_{\mathrm{UK}} & \,. \label{eqn:matheron_prec_form}
\end{align}
We have ready access to efficient evaluation of the sparse $\trans{\mCholPrec}$, as discussed in the Jacobi method. With suitable book-keeping, we can produce the appropriately shuffled local connection filters $\mAllFilterBanks_{\mathrm{shuff}} \leftarrow \textrm{Shuffle}(\mAllFilterBanks)$ and permuted off-diagonal terms $\mOffDiagShuffled \leftarrow \textrm{Shuffle}(\mOffDiag)$ to provide a similar evaluation of the sparse $\mCholPrec$. This product results in a sparse banded diagonal structure in the precision matrix $\mPrecision$. The appropriate blocks of this sparse matrix can be accessed and used to solve for the conditional update of~\eqref{eqn:matheron_rule} using a precision form of the update~\eqref{eqn:matheron_prec_form}.

\subsection{Extension to Non-Gaussian Likelihoods} \label{sec:non-gauss-likelihood}
For many dense prediction tasks, a multivariate Gaussian distribution is not an appropriate likelihood over the observations directly.
However, SUPN is still applicable for this use case, by fitting the multivariate Gaussian distribution to the logit space, \ie to the layer just before the last non-linear layer. This is then followed by an appropriate activation function.
For example, for depth prediction, the outputs of the network should be non-negative and the activation function used is a scaled sigmoid, following monodepth2 \cite{godard2019digging}.
Similarly, for the task of segmentation, the fitting of the SUPN could be done in logit space and soft-max would be used as the activation function.

\subsection{Implementation Details}
\paragraph{Architecture} 
We build upon the U-Net architecture used by Monodepth2 \cite{godard2019digging}, \ie an encoder-decoder architecture where the encoder is a ResNet18 and there are skip connections between the encoder and the decoder.
We add an additional decoder to predict the Cholesky parameters. This decoder takes skip connections from the mean decoder as input. The additional decoder concatenates coordinate maps in the convolutional blocks~\cite{liu2018intriguing} to provide additional spatial information. We designed an off-diagonal prediction approach where the scale of the values is initially very small, $\approx \exp(-4)$, but adapts during training. We found this inductive bias, in lieu of formal priors, was required to predict high quality covariances. 
We use a $5\times5$ neighborhood for the Cholesky decomposition; please see the supplementary details for architecture details and ablation experiments.

\paragraph{Model size}
Our model encodes the distillation of an ensemble of large models into a
single framework; we use only 24\% more parameters than a
single network (out of 8 in the ensemble).

\paragraph{Multi-scale loss}
For depth prediction, we use a multi-scale loss similar to Monodepth2 \cite{godard2019digging}, where the loss in \eqref{eq:loss} is applied across different scales.

\paragraph{Complexity}
Fixed sparsity ensures that all operations are $\mathcal{O}(N)$ for both computation and storage.
Sampling is $\mathcal{O}(J)$ (we used $J = 1000$); 
empirically, the total time for a full Jacobi sample was 0.6s.

\section{Experiments}

For the experiments, we show our method applied to monocular depth estimation.
We use the KITTI dataset~\cite{geiger2013vision} and base our implementation on the Monodepth2 repository~\cite{godard2019digging} and the repository from~\cite{poggi2020uncertainty}.

\paragraph{Pre-trained ensembles}
We use the pre-trained models provided by \cite{poggi2020uncertainty}. In particular, the ensembles created through  bootstrapping together with predictive uncertainty. Two different approaches are used for predictive uncertainty. Both use a diagonal multivariate Laplace distribution, but differ in the way they are trained: \textsc{Log} is trained by directly optimizing the log-likelihood of a self-supervised depth model; while \textsc{Self} uses a pretrained network for depth prediction, without uncertainty estimation, as the teacher model.

\paragraph{Metrics}
For evaluating the accuracy of the estimated depth maps we use a subset of the metrics commonly used for the Kitti dataset: absolute relative error, root mean squared error (RMSE) and the A1 metric.

For evaluation of the uncertainty estimates, we use the metrics used in \cite{poggi2020uncertainty}: area under the sparsification error (AUSE) and area under the random gain (AURG). Both these metrics rely on using per-pixel uncertainty estimates to rank pixels from less confident to more confident. 
For AUSE, this ranking is compared with an oracle ranking that sorts pixels from higher error to lower error, using the different ground truth metrics for ranking. A small AUSE means that the ranking provided by the uncertainty estimate is similar to this oracle ranking.
AURG compares the ranking based on estimated uncertainty with a random ranking, large values are preferred for this metric.

Since both these established metrics only consider per-pixel estimates, we also evaluate the posterior log-likelihood of test samples from the ensembles under our model. To provide a baseline, we also train a version of our model with only a diagonal covariance structure (per-pixel), which cannot model structure. Comparing against this baseline allows us to determine if the model has correctly captured the distribution of test samples and avoided overfitting. We also measure the log-likelihood to other ensembles to ensure that the SUPN variants estimate distributions that generalise well to support other plausible samples.

\begin{table*}[t]
  \caption{\textbf{Accuracy comparison}: Quantitative comparison of quality on 
  commonly used depth metrics (see supplement for the remaining metrics in \cite{godard2019digging}). The ``Best'' metrics sample 40 different 
  predictions for our model, and from the 8 ensembles for the baseline, and choose the best under each metric. Standard deviations are given in brackets. The box plot illustrates the substantial overlap in distributions.}
  \label{tab:accuracy_unconditioned}
  \centering
  \begin{minipage}{0.23\textwidth}
  \includegraphics[width=\textwidth]{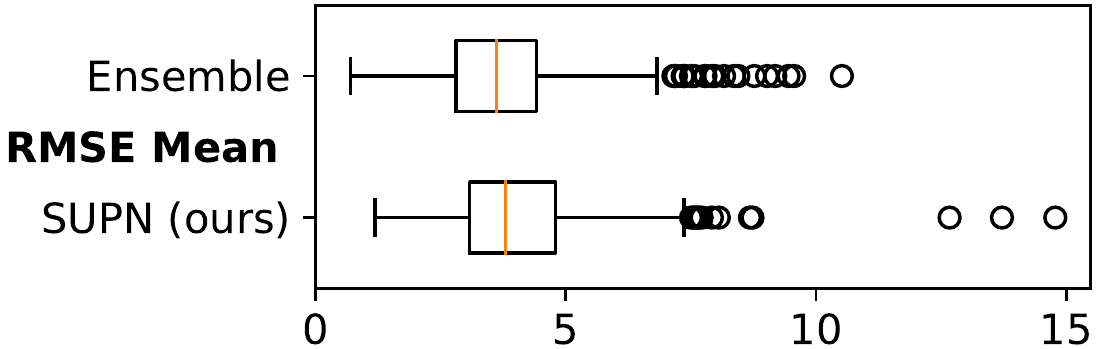}\\
  \footnotesize Box plot illustrating the strong distribution overlap between the original ensemble and the trained SUPN model for Boot+Log RMSE mean.
  \end{minipage}\hfill
  \begin{adjustbox}{max width=0.75\textwidth}
  \begin{tabular}{lrrrrrr}
    \toprule
    Model name     &  AbsRel Mean $\downarrow$&   AbsRel Best $\downarrow$ &  RMSE Mean $\downarrow$ & RMSE Best $\downarrow$ & A1 Mean $\uparrow$ & A1 Best $\uparrow$ \\
    \midrule
    MD2 Boot+Log & 0.092 (0.035) & 0.084 (0.031) & 3.850 (1.370) & 3.600 (1.260) & 0.911 (0.064) & 0.923 (0.055) \\
    MD2 Boot+Self & 0.088 (0.034) & 0.083 (0.031) & 3.795 (1.397) & 3.574 (1.323) & 0.918 (0.060) & 0.929 (0.051)\\
    \midrule
    Diagonal  & 0.101 (0.044) & 0.103 (0.043) & 4.000 (1.457) & 4.020 (1.444) & 0.896 (0.076) & 0.894 (0.074) \\ 
    SUPN Boot+Log & 0.104 (0.047) & 0.095 (0.039) & 4.071 (1.489) & 3.577 (1.191) & 0.892 (0.080)  & 0.909 (0.069) \\
    SUPN Boot+Self & 0.103 (0.049) & 0.096 (0.046) & 4.091 (1.442) & 3.800 (1.396) & 0.894 (0.078) & 0.906 (0.073) \\
    \bottomrule
  \end{tabular}
  \end{adjustbox}
\end{table*}

\begin{table*}[t]
  \caption{\textbf{Pixelwise uncertainty metrics}: AUSE (area under the sparsification error), lower is better. AURG (area under the random gain), higher is better. Uncertainy for SUPN estimated from std-deviation of 10 samples. Results marked with a * differ from the published work by \cite{poggi2020uncertainty}, as to make it comparable we do not use the Monodepth 1 post-processing. LL (Log-Likelihood) columns provide the log-likelihood of samples from the respective ensembles under the diagonal (baseline) and SUPN models. Standard deviations are given in brackets.}
  \label{tab:accuracy_uncertainty}
  \centering
  \begin{adjustbox}{max width=\textwidth}
  \newcommand{\lloffset}[0]{$\times10^{5}$}
  \begin{tabular}{lrrrrrrrr}
    \toprule
    Model name     &  AbsRel AUSE $\downarrow$  &   AbsRel AURG $\uparrow$  &  RMSE AUSE $\downarrow$ & RMSE AURG $\uparrow$  & A1 AUSE $\downarrow$ & A1 AURG $\uparrow$ & LL Boot+Log \lloffset $\uparrow$ & LL Boot+Self \lloffset $\uparrow$ \\
    \midrule
    MD2 Boot+Log & 0.038 (0.020)  & 0.021 (0.019) & 2.449 (0.877) & 0.820 (0.929) & 0.046 (0.048) & 0.037 (0.040) & & \\
    MD2 Boot+Self & 0.029 (0.018) & 0.028 (0.019) & 1.924 (1.006) & 1.316 (1.000) & 0.028 (0.041) & 0.049 (0.037) & & \\
    \midrule
    MD2 Boot+Log* & 0.041 (0.019)  & 0.018 (0.020) & 2.927 (1.327) & 0.324 (1.019) & 0.050 (0.049) & 0.032 (0.037) \\
    MD2 Boot+Self* & 0.040 (0.021) & 0.017 (0.018) & 2.906 (1.458) & 0.331 (1.08) & 0.045 (0.045) & 0.031 (0.035) & & \\
    \midrule
    Diagonal & 0.085 (0.050) & -0.020 (0.030) & 5.075 (1.924) & -1.697 (0.799) & 0.138 (0.083) & -0.440 (0.053) &  1.77 (11.48) & 1.15 (12.78) \\
    SUPN Boot+Log &  0.037 (0.027) & 0.030 (0.025) & 1.555 (1.307) & 1.856 (1.355) & 0.040 (0.063) & 0.058 (0.047)  & 40.60 (1.35) & 38.18  (2.93)\\
    SUPN Boot+Self & 0.050 (0.037) & 0.017 (0.028) & 2.786 (1.796) & 0.674 (1.544) & 0.062 (0.074) & 0.034 (0.055)  &  36.51 (2.31) & 38.87 (1.63)\\
    \bottomrule
  \end{tabular}
  \end{adjustbox}
\end{table*}

\subsection{Quantitative Results}

\paragraph{Depth accuracy}
In Table \ref{tab:accuracy_unconditioned} we show a quantitative comparison between the two variants of ensembles and the corresponding versions of our model, trained to approximate them. We compare the methods using the depth estimation metrics.
While the mean performance of the ensembles is slightly superior to our approximate models, the results are comparable within the margin of error. The box plot in Table~\ref{tab:accuracy_unconditioned} highlights the strong overlap in the error distribution of the ensemble and SUPN models, indicating that despite the significant reduction in the number of parameters, SUPN is able to approximate the performance of the ensemble.

Our models compare favourably with a diagonal only model.
This is particularly noticeable in the metrics for the best sample. Samples from our model consistently outperform samples from a diagonal only Gaussian.

\begin{figure*}[ht!]
    \centering
    \renewcommand{\arraystretch}{0.3}
    \begin{adjustbox}{max width=0.85\textwidth}
    \begin{tabular}{ccc}
        \includegraphics[width=0.3\textwidth]{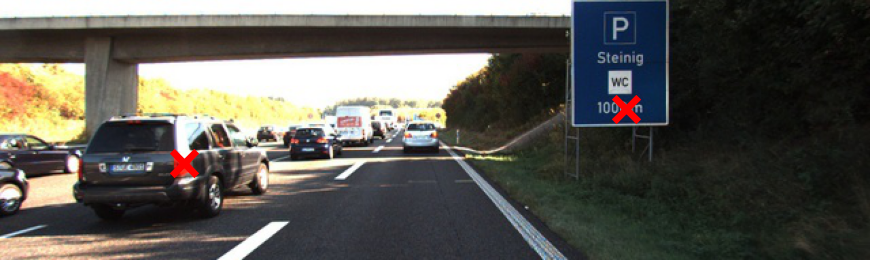}
        &
        \includegraphics[width=0.3\textwidth]{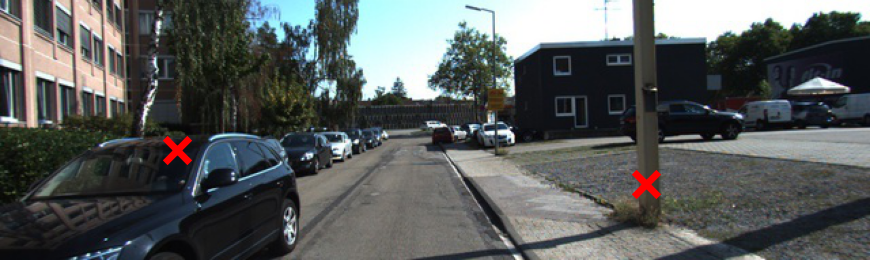}
        &
        \includegraphics[width=0.3\textwidth]{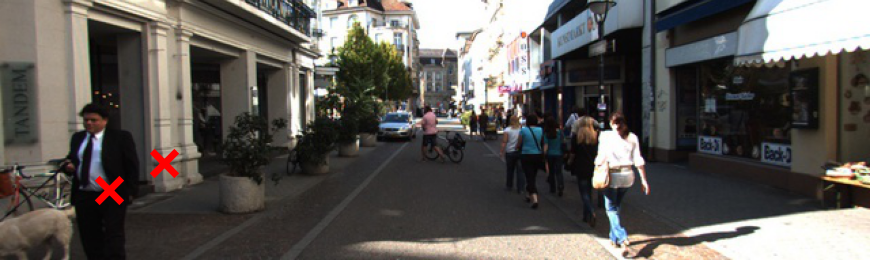}
        \\
        \includegraphics[width=0.3\textwidth]{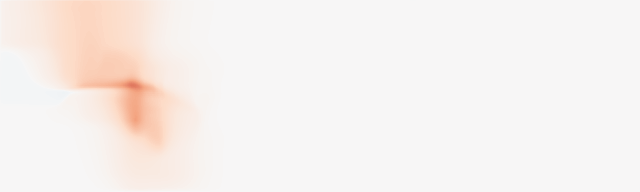}
        &
        \includegraphics[width=0.3\textwidth]{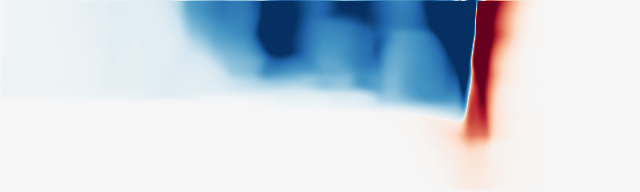}
        &
        \includegraphics[width=0.3\textwidth]{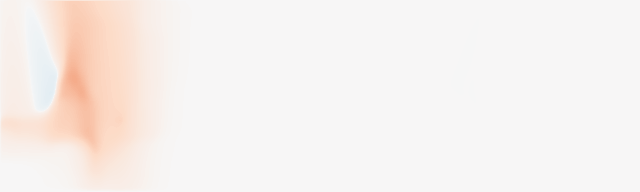}
        \\
        \includegraphics[width=0.3\textwidth]{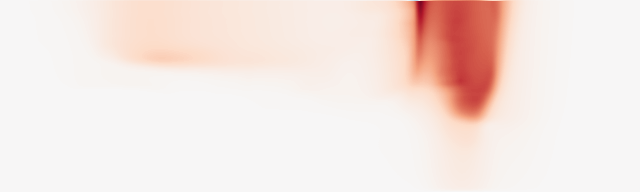}
        &
        \includegraphics[width=0.3\textwidth]{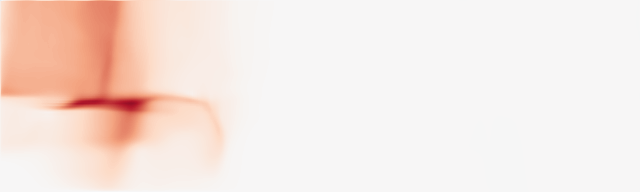}
        &
        \includegraphics[width=0.3\textwidth]{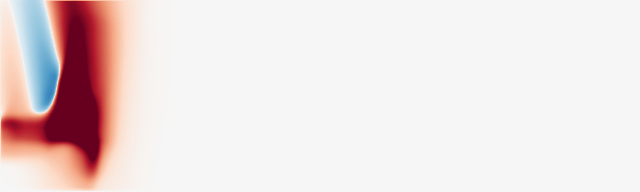}
    \end{tabular}
    \end{adjustbox}
    \caption{Visualisation of learned covariances between example pixels (red crosses) and other pixel locations for SUPN \textsc{Boot+Log}. Red indicates high positive correlation, blue is strong negative correlation. 
    For clarity, these plots are scaled into the standard deviation range (via a signed square root operation) and plotted over a range [-0.05, 0.05].
    These examples illustrate the long range correlations that can be captured from very local structure ($5 \times 5$ pixel regions) in the precision matrix. For more examples, see the supplementary video.
    }\label{fig:cov}
\end{figure*}

\paragraph{Uncertainty estimation}
Table~\ref{tab:accuracy_uncertainty} provides a quantitative comparison in terms of uncertainty metrics. SUPN consistently outperforms the teacher ensemble model for both \textsc{Log} and \textsc{Self}. 
The log-likehood values demonstrate that the correlated structure capture by SUPN is better able to explain the test outputs of the ensembles that the baseline diagonal model. 
The samples routinely have higher support under the SUPN model which suggests that some of the other measures are not accurately measure the quality of the structure present in the posterior predictions of the model.
As expected, the performance of the SUPN approaches on the test set for the corresponding samples are slightly better but we note that overall the values are similar between the two methods indicating that the correlations captured are not overfit to the specific ensembles.

\subsection{Qualitative Results}

\begin{figure*}[t]
\centering
\includegraphics[width=\textwidth]{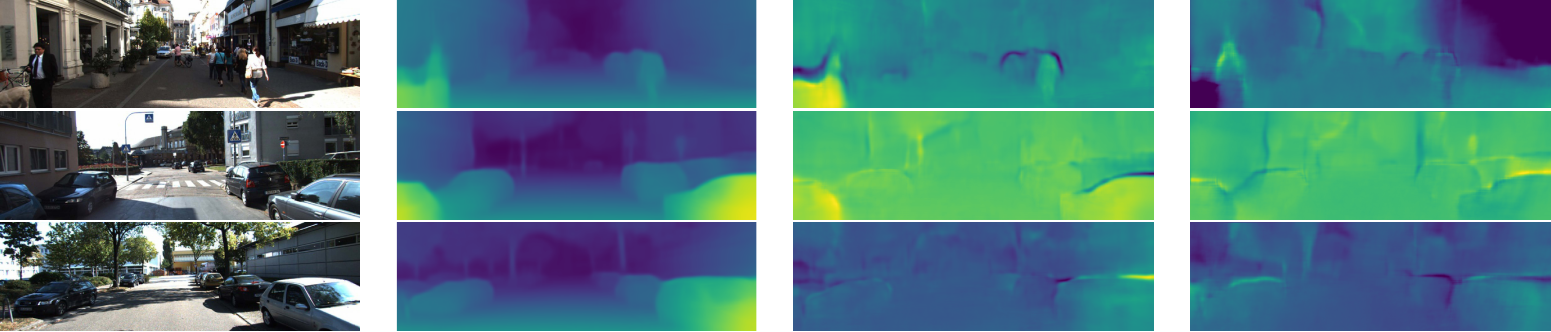}\\
\renewcommand{\arraystretch}{0.3}
\begin{adjustbox}{max width=\textwidth}
\begin{tabular}{CCCC}
    (a) Input image & (b) Mean disparity & (c) \textsc{Boot+Log} diff  &  (d) SUPN \textsc{Boot+Log} diff
\end{tabular}
\end{adjustbox}
\caption{Example depth samples (see supplementary video for more).  (b)~Average normalised disparity predicted by the ensemble models. Difference between the mean and one of the (c)~\textsc{Boot+Log} ensemble or (d)~SUPN \textsc{Boot+Log}; the samples appear qualitatively similar.} \label{fig:samples}
\end{figure*}
\begin{figure*}[h!]
\centering
\renewcommand{\arraystretch}{0.3}
\begin{adjustbox}{max width=0.63\textwidth}
\begin{tabular}{ccc}
    \includegraphics[width=0.58\textwidth]{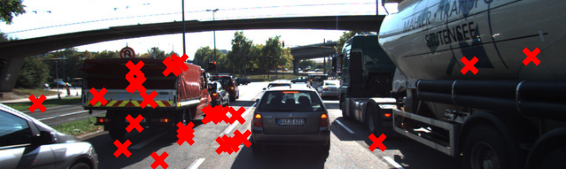}
    &
    \includegraphics[width=0.58\textwidth]{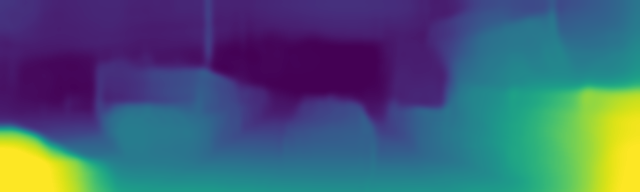}
    &
    \includegraphics[width=0.58\textwidth]{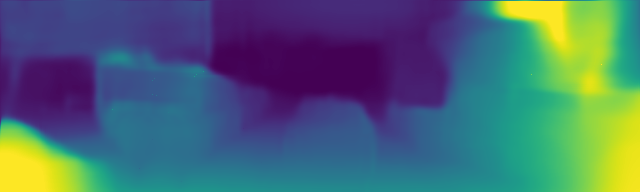}
    \\
    \includegraphics[width=0.58\textwidth]{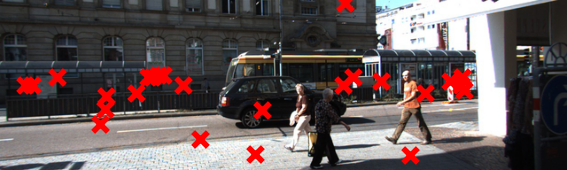}
    &
    \includegraphics[width=0.58\textwidth]{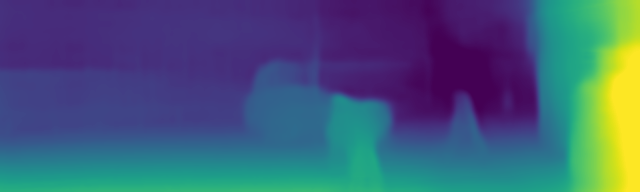}
    &
    \includegraphics[width=0.58\textwidth]{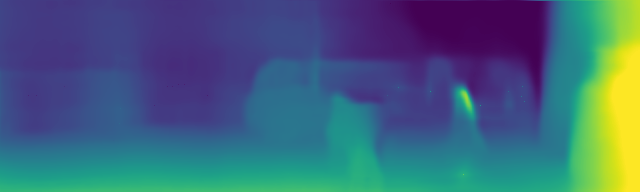}
    \\
    \includegraphics[width=0.58\textwidth]{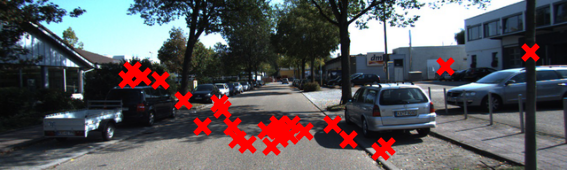}
    &    
    \includegraphics[width=0.58\textwidth]{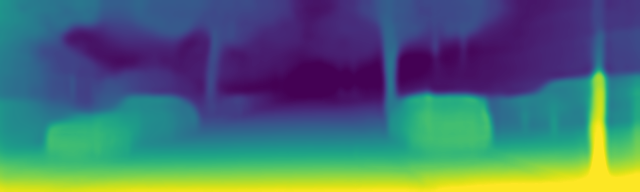}
    &
    \includegraphics[width=0.58\textwidth]{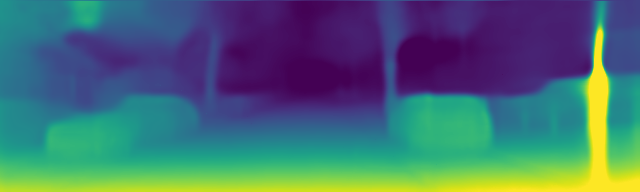}
    \\[5pt]
    \huge
    (a) Image + conditioning locations & \huge (b) Mean disparity & \huge (c) Conditional mean disparity
\end{tabular}
\end{adjustbox}
\begin{adjustbox}{max width=0.35\textwidth}
\begin{tabular}{c}
  \begin{tabular}{lrrr}
    \toprule
    Model name     &  AbsRel  $\downarrow$&   RMSE $\downarrow$ & A1 $\uparrow$ \\
    \midrule
    Ensemble & 0.088 (0.034) & 3.795 (1.397) & 0.918 (0.060) \\
    \midrule
    SUPN & 0.104 (0.047) & 4.071 (1.489)  & 0.892 (0.080) \\
    \midrule
    SUPN 25$^*$  & 0.086 (0.036) &3.893 (1.541) & 0.922 (0.053)  \\
    SUPN 50$^*$ & 0.076 (0.031)  &3.607 (1.433) &0.936 (0.041) \\
    SUPN 100$^*$ & 0.064 (0.028) & 3.251 (1.290) &0.949 (0.033) \\
    SUPN 200$^*$ & 0.054 (0.022)  &2.873 (1.100) &0.960 (0.028) \\
    \bottomrule
  \end{tabular} 
  \\[102pt]
    \large (d) Quantitative conditioning accuracy
  \end{tabular}
  \end{adjustbox}

\caption{Conditional prediction using sparse ground truth depth information. (a) shows 25 randomly sampled conditioning locations that have valid depth (b) shows the original mean disparity, while (c) shows the conditional mean disparity. d) Quantifies the accuracy improvement of our SUPN Boot+Log model when conditioned on $N^*$ random ground truth depth pixels, repeated 10 times per image. } \label{fig:cond_dist}
\end{figure*}

Figure~\ref{fig:samples} illustrates samples from the \textsc{Boot+Log} ensemble, and the SUPN approximation; the samples are visually similar and exhibit considerable long-range structure. 

\paragraph{Introspection}
As discussed in \S~\ref{sec:introduction}, one of the advantages of an explicit distribution is that it allows for introspection. Figure~\ref{fig:cov} illustrates how the covariance between a specified pixel, and any other, can be explicitly computed. 
These visualisations are the corresponding row of the covariance matrix obtained using the sampling process of~\eqref{eqn:sampling_with_precision} twice (with $\mCholPrec$ and then $\trans{\mCholPrec}$) to a one-hot vector encoding the pixel of interest on the RHS (instead of $\mRawSamples$) and no mean.

\paragraph{Conditional Distributions}
Figure~\ref{fig:cond_dist} illustrates our model's ability to condition samples on arbitrary output pixels, which is not possible in most deep probabilistic models. In this example, we use some samples from the ground truth depth as additional conditioning on the predicted distribution, and show the conditional mean.

\section{Discussion and Limitations}
Similar to other distillation methods, the performance of our method is upper-bounded by the performance of the original ensemble model. This might become an issue when the ensemble is small, and doesn't capture the full diversity of data. 
We've observed visually that the log-likelihood alone is not always a good predictor of sample quality, \ie a sample might have high log-likelihood while looking implausible, this may be due to a lack of variation in the ensemble predictions. This could potentially be overcome by using priors on the predicted Gaussian distribution, and future work will consider subsequently training the model on the specified task, using the drawn samples. 

As a deterministic approximation to the output of an ensemble, we seek to capture all forms of uncertainty captured by the ensemble (e.g.~aleatoric and epistemic). We acknowledge that we do not consider the epistemic uncertainty in the approximation separately, however our work may be considered orthogonal to work in this area, e.g.~BNNs, and could be readily combined.

\paragraph{Potential negative impact}
We think that uncertainty estimation is a valuable endeavor in improving deep models, and that our approach of using explicit distributions is a step in the right direction, providing tangible benefits. However, the predicted distributions have yet to be evaluated on out-of-distribution data. As with most machine learning models, we cannot expect generalisation of our SUPN prediction networks to very different data. Clearly, this approach will require extensive validation before deployment in safety-critical applications, such as autonomous driving.

\paragraph{Conclusion}
We presented a method for uncertainty estimation by distillation of ensemble models. We showed that our structured Gaussian model can be predicted by a single pass of a convolutional neural network, we have proposed an efficient method for drawing samples.

Our method was validated on the task of depth prediction from a single image. Our distilled model is able to perform similarly to the original ensemble on uncertainty metrics, while requiring fewer parameters and allowing arbitrary numbers of samples to be drawn. We have illustrated that the samples capture long-range correlations in the image domain, which is in stark contrast to prior works that use diagonal covariance matrices.
We demonstrated the benefit of our predicted distribution in terms of enabling arbitrary test-time conditioning and allowing for direct introspection of the inferred distribution.
We hope that our paper sparks interest in predictive uncertainty models that are able to model correlation in the output space, with many practical applications in computer vision and integration with subsequent down-stream tasks. 
%

{\small
\bibliographystyle{ieee_fullname}
\bibliography{structured_seg_uncertainty.bib}

\begin{thebibliography}{10}\itemsep=-1pt

\bibitem{abdar2020review}
Moloud Abdar, Farhad Pourpanah, Sadiq Hussain, Dana Rezazadegan, Li Liu,
  Mohammad Ghavamzadeh, Paul Fieguth, Abbas Khosravi, U~Rajendra Acharya,
  Vladimir Makarenkov, et~al.
\newblock A review of uncertainty quantification in deep learning: Techniques,
  applications and challenges.
\newblock {\em arXiv preprint arXiv:2011.06225}, 2020.

\bibitem{bulo2016dropout}
Samuel~Rota Bul{\`o}, Lorenzo Porzi, and Peter Kontschieder.
\newblock Dropout distillation.
\newblock In {\em International Conference on Machine Learning}, pages 99--107.
  PMLR, 2016.

\bibitem{chandra2016fast}
Siddhartha Chandra and Iasonas Kokkinos.
\newblock Fast, exact and multi-scale inference for semantic image segmentation
  with deep gaussian crfs.
\newblock In {\em European conference on computer vision}, pages 402--418.
  Springer, 2016.

\bibitem{dorta2018structured}
Garoe Dorta, Sara Vicente, Lourdes Agapito, Neill~DF Campbell, and Ivor
  Simpson.
\newblock Structured uncertainty prediction networks.
\newblock In {\em Proceedings of the IEEE Conference on Computer Vision and
  Pattern Recognition}, pages 5477--5485, 2018.

\bibitem{gal2016dropout}
Yarin Gal and Zoubin Ghahramani.
\newblock Dropout as a bayesian approximation: Representing model uncertainty
  in deep learning.
\newblock In {\em international conference on machine learning}, pages
  1050--1059. PMLR, 2016.

\bibitem{geiger2013vision}
Andreas Geiger, Philip Lenz, Christoph Stiller, and Raquel Urtasun.
\newblock Vision meets robotics: The kitti dataset.
\newblock {\em The International Journal of Robotics Research},
  32(11):1231--1237, 2013.

\bibitem{godard2017monodepth}
Cl{\'e}ment Godard, Oisin Mac~Aodha, and Gabriel~J Brostow.
\newblock Unsupervised monocular depth estimation with left-right consistency.
\newblock In {\em Proceedings of the IEEE conference on computer vision and
  pattern recognition}, pages 270--279, 2017.

\bibitem{godard2019digging}
Cl{\'e}ment Godard, Oisin Mac~Aodha, Michael Firman, and Gabriel~J Brostow.
\newblock Digging into self-supervised monocular depth estimation.
\newblock In {\em Proceedings of the IEEE/CVF International Conference on
  Computer Vision}, pages 3828--3838, 2019.

\bibitem{huang2017snapshot}
Gao Huang, Yixuan Li, Geoff Pleiss, Zhuang Liu, John~E Hopcroft, and Kilian~Q
  Weinberger.
\newblock Snapshot ensembles: Train 1, get m for free.
\newblock In {\em ICLR}, 2017.

\bibitem{rtf2012}
Jeremy Jancsary, Sebastian Nowozin, Toby Sharp, and Carsten Rother.
\newblock Regression tree fields: An efficient, non-parametric approach to
  image labeling problems.
\newblock In {\em Proceedings of the IEEE conference on computer vision and
  pattern recognition}, 2012.

\bibitem{kendall2017uncertainties}
Alex Kendall and Yarin Gal.
\newblock What uncertainties do we need in bayesian deep learning for computer
  vision?
\newblock In {\em Advances in neural information processing systems}, pages
  5574--5584, 2017.

\bibitem{kingma2015adam}
Diederik~P. Kingma and Jimmy Ba.
\newblock Adam: {A} method for stochastic optimization.
\newblock In {\em ICLR}, 2015.

\bibitem{kingma2013auto}
Diederik~P Kingma and Max Welling.
\newblock Auto-encoding variational bayes.
\newblock In {\em ICLR}, 2014.

\bibitem{klodt2018supervising}
Maria Klodt and Andrea Vedaldi.
\newblock Supervising the new with the old: learning sfm from sfm.
\newblock In {\em Proceedings of the European Conference on Computer Vision
  (ECCV)}, pages 698--713, 2018.

\bibitem{lakshminarayanan2017simple}
Balaji Lakshminarayanan, Alexander Pritzel, and Charles Blundell.
\newblock Simple and scalable predictive uncertainty estimation using deep
  ensembles.
\newblock {\em NeurIPS}, 2017.

\bibitem{li2020improving}
Zhizhong Li and Derek Hoiem.
\newblock Improving confidence estimates for unfamiliar examples.
\newblock In {\em CVPR}, 2020.

\bibitem{liu2018intriguing}
Rosanne Liu, Joel Lehman, Piero Molino, Felipe~Petroski Such, Eric Frank, Alex
  Sergeev, and Jason Yosinski.
\newblock An intriguing failing of convolutional neural networks and the
  coordconv solution.
\newblock {\em arXiv preprint arXiv:1807.03247}, 2018.

\bibitem{mackay1995probable}
David~JC MacKay.
\newblock Probable networks and plausible predictions-a review of practical
  bayesian methods for supervised neural networks.
\newblock {\em Network: computation in neural systems}, 1995.

\bibitem{malinin2020ensemble}
Andrey Malinin, Bruno Mlodozeniec, and Mark Gales.
\newblock Ensemble distribution distillation.
\newblock {\em ICLR}, 2020.

\bibitem{poggi2020uncertainty}
Matteo Poggi, Filippo Aleotti, Fabio Tosi, and Stefano Mattoccia.
\newblock On the uncertainty of self-supervised monocular depth estimation.
\newblock In {\em Proceedings of the IEEE/CVF Conference on Computer Vision and
  Pattern Recognition}, pages 3227--3237, 2020.

\bibitem{prince2012computer}
Simon~JD Prince.
\newblock {\em Computer vision: models, learning, and inference}.
\newblock Cambridge University Press, 2012.

\bibitem{shen2021real}
Yichen Shen, Zhilu Zhang, Mert~R Sabuncu, and Lin Sun.
\newblock Real-time uncertainty estimation in computer vision via
  uncertainty-aware distribution distillation.
\newblock In {\em Proceedings of the IEEE/CVF Winter Conference on Applications
  of Computer Vision}, pages 707--716, 2021.

\bibitem{srivastava2014dropout}
Nitish Srivastava, Geoffrey Hinton, Alex Krizhevsky, Ilya Sutskever, and Ruslan
  Salakhutdinov.
\newblock Dropout: a simple way to prevent neural networks from overfitting.
\newblock {\em The journal of machine learning research}, 2014.

\bibitem{vemulapalli2016gaussian}
Raviteja Vemulapalli, Oncel Tuzel, Ming-Yu Liu, and Rama Chellapa.
\newblock Gaussian conditional random field network for semantic segmentation.
\newblock In {\em Proceedings of the IEEE conference on computer vision and
  pattern recognition}, pages 3224--3233, 2016.

\bibitem{williams1996using}
Peter~M Williams.
\newblock Using neural networks to model conditional multivariate densities.
\newblock {\em Neural computation}, 8(4):843--854, 1996.

\bibitem{xia2020generating}
Zhihao Xia, Patrick Sullivan, and Ayan Chakrabarti.
\newblock Generating and exploiting probabilistic monocular depth estimates.
\newblock In {\em Proceedings of the IEEE/CVF Conference on Computer Vision and
  Pattern Recognition}, pages 65--74, 2020.

\end{thebibliography}
}
\newpage
\onecolumn
\appendix
\section{Sample and covariance videos}
To demonstrate the variability we capture in our predicted covariance matrix, we provide videos containing the 8 samples from the ensembles (on repeat) and 100 samples from our approach. As can be seen, we capture similar modes of variability, but with the added benefit of being able to generate more samples.

To enable detailed illustration of the covariance between pixels, we provide videos showing the covariance between highlighted pixels, and all other pixels in the image. These videos allow for a broader view of the patterns of correlation that are learned by our model.

\section{Network Architecture}

\begin{figure*}[!hbt]
    \centering
    \includegraphics[width=0.9\textwidth]{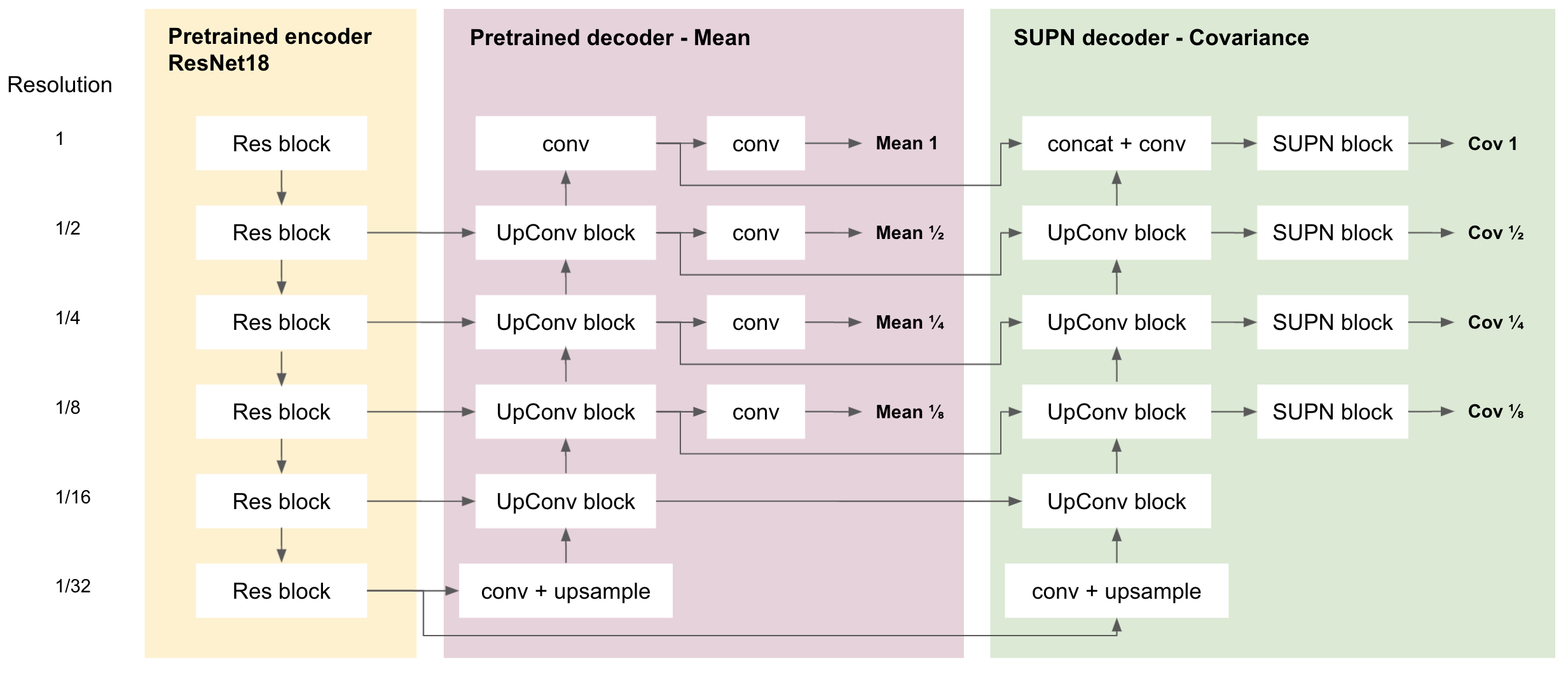}
    \caption{\textbf{Network architecture} }
    \label{fig:network_architecture}
\end{figure*}

\begin{figure*}[!hbt]
    \centering
    \includegraphics[width=0.8\textwidth]{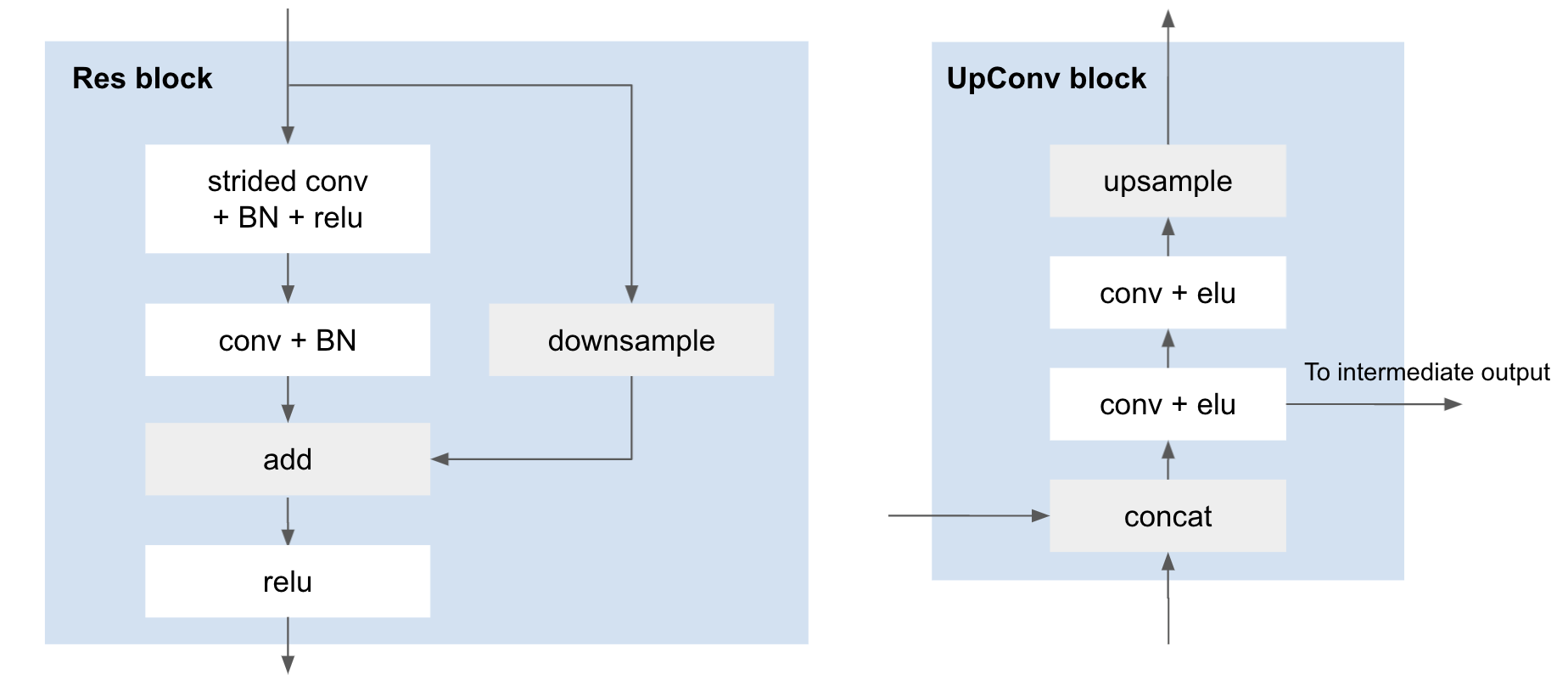}
    \caption{\textbf{Detail view of Res block and UpConv block}}
    \label{fig:resbloc_upconvblock}
\end{figure*}

\begin{figure*}[!hbt]
    \centering
    \includegraphics[width=0.8\textwidth]{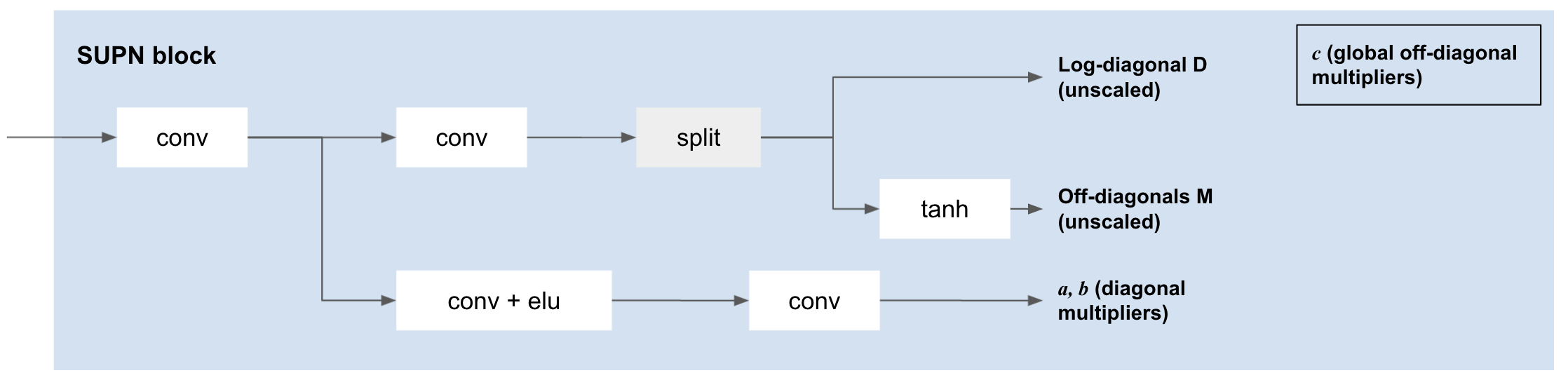}
    \caption{\textbf{Detail view of SUPN block}}
    \label{fig:supn_block}
\end{figure*}

Illustrations of our network architecture are provided in figures \ref{fig:network_architecture}, \ref{fig:resbloc_upconvblock} and \ref{fig:supn_block}.

As shown in Fig.~\ref{fig:supn_block}, our SUPN block predicts not only the diagonal and off-diagonal maps, but also scaling factors for those. While the scaling factors for the diagonal terms are image dependent, the scaling factors for off-diagonal elements are shared between all images.

\paragraph{Diagonal scaling}
The final diagonal value is given by:
$\exp(\text{D})\times \exp(a) + \exp(b)$, where $\text{D}$ is the log-diagonal output of the network and $a$ and $b$ are the diagonal multipliers.

\paragraph{Off-diagonal scaling}
The final value for the off-diagonals is given by: $\text{M} \times c$, where $M$ is the output of the network after the non-linearity $tanh$ and $c$ takes a different value for each of the off-diagonal maps corresponding to a different neighbour.

\section{Ablation Experiments}
The following tables extend those in the main submission by removing aspects of the model architecture, or testing model variants. All SUPN variants were trained using the Monodepth2 Boot+Log ensembles from \cite{poggi2020uncertainty}, except SUPN Boot+self. The various SUPN variants are given in Table \ref{table:variants}. 

Accuracy measures for different variants and baselines are given in Table \ref{tab:accuracy_unconditioned_ablation} and uncertainty measures are given in Table \ref{tab:accuracy_uncertainty_ablation}. Key observation include:
\begin{itemize}
    \item Removing the off-diagonal scaling mechanism described above leads to better log-likelihoods, but substantially worse samples and uncertainty metrics. We believe this is because our formulation provides an initial bias towards preferring smaller off-diagonal values, although the magnitude of these can grow this has to be sufficiently supported by an increase in likelihood. 
    \item Using a $3 \times 3$ neighbourhood for the Cholesky of the precision matrix prediction leads to a small reduction in performance.
    \item The use of the additional diagonal scaling branch and concatenated pixel maps lead to reasonably small improvements in mean accuracy and uncertainty metrics, and a reduction in std-deviation for some metrics.
\end{itemize}

\begin{table*}[h]
\caption{SUPN variants for ablation study}\label{table:variants}
\centering
  \begin{adjustbox}{max width=\textwidth}
  \begin{tabular}{ll}
  \toprule
  Model suffix & Difference \\
  \midrule
  $3\times3$ & Uses a $3 \times 3$ pixel neighbourhood rather than $5 \times 5$ \\
  - ODS & No off-diagonal scaling function \\
  - PM & No concatenated pixel coordinate map \\
  - DS & No extra diagonal scaling branch \\
  \bottomrule
  \end{tabular}
  \end{adjustbox}
\end{table*}

\begin{landscape}

\begin{table*}[t]
  \caption{\textbf{Accuracy comparison}: Quantitative comparison of depth quality on three commonly used depth metrics. For the ``Best'' metrics, we sample 40 different depth map predictions for our model, and from the 8 ensembles for the baseline, and choose the best of them according to each metric. Standard deviations are given in brackets.}
  \label{tab:accuracy_unconditioned_ablation}
  \centering
  \begin{adjustbox}{max width=\linewidth}
  \begin{tabular}{lrrrrrr}
    \toprule
    Model name     &  AbsRel Mean $\downarrow$&   AbsRel Best $\downarrow$ &  RMSE Mean $\downarrow$ & RMSE Best $\downarrow$ & A1 Mean $\uparrow$ & A1 Best $\uparrow$ \\
    \midrule
    MD2 Boot+Log & 0.092 (0.035) & 0.084 (0.031) & 3.850 (1.370) & 3.600 (1.260) & 0.911 (0.064) & 0.923 (0.055) \\
    MD2 Boot+Self & \textbf{0.088} (0.034) & \textbf{0.083} (0.031) & \textbf{3.795} (1.397) & \textbf{3.574} (1.323) & \textbf{0.918} (0.060) & \textbf{0.929} (0.051)\\
    \midrule
    Diagonal  & 0.101 (0.44) & 0.103 (0.43) & 4.00 (1.457) & 4.02 (1.444) & 0.896 (0.076) & 0.894 (0.074) \\
    SUPN $3\times3$ & 0.104 (0.047) & 0.100 (0.044) &4.088 (1.510) & 3.876 (1.422) &0.893 (0.079) & 0.902 (0.073)  \\
    SUPN - ODS & 0.107 ( 0.048) & 0.103 ( 0.048) &4.148 ( 1.514) & 3.965 ( 1.625) &0.887 ( 0.081) & 0.899 ( 0.074) \\
    SUPN - PM & 0.104 (0.046) & 0.096 (0.039) &4.077 (1.491) & 3.663 (1.261) &0.892 (0.079) & 0.908 (0.069) \\
    SUPN - DS  & 0.104 (0.046) & 0.097 (0.041) &4.072 (1.502) & 3.700 (1.304) &0.892 (0.078) & 0.908 (0.069) \\
    SUPN Boot+Log & 0.104 (0.047) & 0.095 (0.039) & 4.071 (1.489) & 3.577 (1.191) & 0.892 (0.080)  & 0.909 (0.069) \\
    SUPN Boot+Self & 0.103 (0.049) & 0.096 (0.046) & 4.091 (1.442) & 3.800 (1.396) & 0.894 (0.078) & 0.906 (0.073) \\
    \bottomrule
  \end{tabular}
  \end{adjustbox}
\end{table*}

\begin{table*}[t]
  \caption{\textbf{Pixelwise uncertainty metrics}: AUSE (area under the sparsification error), lower is better. AURG (area under the random gain), higher is better. Uncertainy for SUPN estimated from std-deviation of 10 samples. Results marked with a * differ from the published work by \cite{poggi2020uncertainty}, as to make it comparable we do not use the Monodepth 1 post-processing. LL (Log-Likelihood) columns provide the log-likelihood of samples from the respective ensembles under the diagonal (baseline) and SUPN models. Standard deviations are given in brackets.}
  \label{tab:accuracy_uncertainty_ablation}
  \centering
  \begin{adjustbox}{max width=\linewidth}
  \newcommand{\lloffset}[0]{$\times10^{5}$}
  \begin{tabular}{lrrrrrrrr}
    \toprule
    Model name     &  AbsRel AUSE $\downarrow$  &   AbsRel AURG $\uparrow$  &  RMSE AUSE $\downarrow$ & RMSE AURG $\uparrow$  & A1 AUSE $\downarrow$ & A1 AURG $\uparrow$ & LL Boot+Log \lloffset $\uparrow$ & LL Boot+Self \lloffset $\uparrow$ \\
    \midrule
    MD2 Boot+Log & 0.038 (0.020)  & 0.021 (0.019) & 2.449 (0.877) & 0.820 (0.929) & 0.046 (0.048) & 0.037 (0.040) & & \\
    MD2 Boot+Self & \textbf{0.029} (0.018) & 0.028 (0.019) & 1.924 (1.006) & 1.316 (1.000) & \textbf{0.028} (0.041) & 0.049 (0.037) & & \\
    \midrule
    MD2 Boot+Log* & 0.041 (0.019)  & 0.018 (0.020) & 2.927 (1.327) & 0.324 (1.019) & 0.050 (0.049) & 0.032 (0.037) \\
    MD2 Boot+Self* & 0.040 (0.021) & 0.017 (0.018) & 2.906 (1.458) & 0.331 (1.08) & 0.045 (0.045) & 0.031 (0.035) & & \\
    \midrule
    Diagonal & 0.085 (0.050) & -0.020 (0.030) & 5.075 (1.924) & -1.697 (0.799) & 0.138 (0.083) & -0.440 (0.053) &  1.77 (11.48) & 1.15 (12.78 ) \\
    SUPN $3\times3$ & 0.037 ( 0.028) & \textbf{0.030} ( 0.027) &1.922 ( 1.472) & 1.525 ( 1.450) &0.041 ( 0.062) & 0.056 ( 0.049) & 38.12 (1.85)& 35.95 (2.61) \\
    SUPN - ODS & 0.060 ( 0.040) & 0.009 ( 0.027) &3.359 ( 1.788) & 0.130 ( 1.249) &0.086 ( 0.081) & 0.016 ( 0.053) & \textbf{41.92} (2.40) & 38.28 (3.93) \\
    SUPN - PM & 0.039 ( 0.026) & 0.028 ( 0.025) &1.853 ( 1.426) & 1.583 ( 1.401) &0.044 ( 0.062) & 0.053 ( 0.048) & 40.54 (1.39) & 38.18 (2.27)  \\
    SUPN - DS  & 0.039 ( 0.027) & 0.027 ( 0.026) &1.785 ( 1.570) & 1.648 ( 1.572) &0.045 ( 0.064) & 0.053 ( 0.050) & 40.54 (1.26) & 38.14 (2.21) \\
    SUPN Boot+Log &  0.037 (0.027) & \textbf{0.030} (0.025) & \textbf{1.555} (1.307) & \textbf{1.856} (1.355) & 0.040 (0.063) & \textbf{0.058} (0.047)  & 40.60 (1.35) & 38.18  (2.93)\\
    SUPN Boot+Self & 0.050 (0.037) & 0.017 (0.028) & 2.786 (1.796) & 0.674 (1.544) & 0.062 (0.074) & 0.034 (0.055)  &  36.51 (2.31) & \textbf{38.87} (1.63)\\
    \bottomrule
  \end{tabular}
  \end{adjustbox}
\end{table*}

\end{landscape}

\section{Sparsity pattern of the Cholesky decomposition}
We include an illustration of the sparsity pattern of the Cholesky decomposition explained in section 2.2 of the paper.

\begin{figure*}[h]
    \centering
    \includegraphics[width=0.7\textwidth]{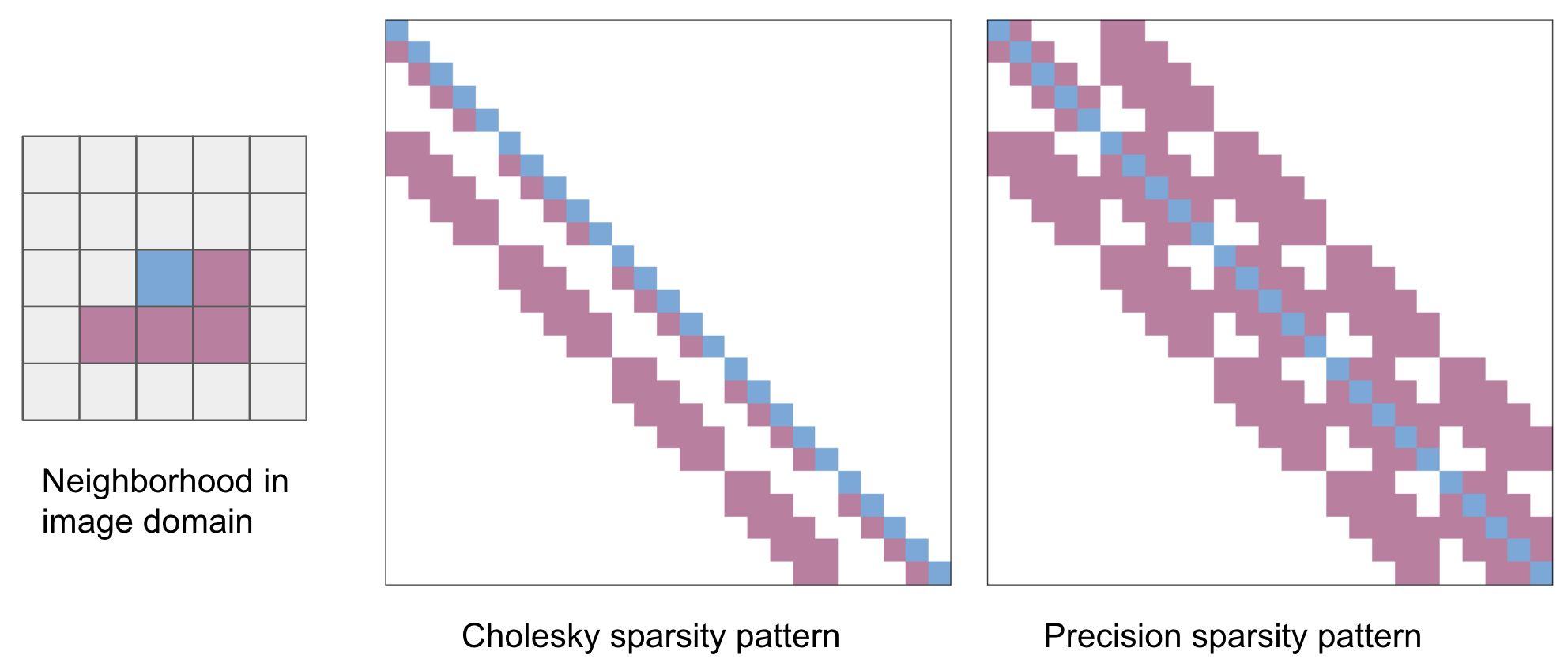}
    \caption{\textbf{Sparsity pattern for the Cholesky decomposition.} (Left)~A $3 \times 3$ neighborhood around a central pixel in blue. Only neighbours in pink are considered in the Cholesky matrix, which ensures that the matrix is sparse and lower triangular. (Center)~The sparsity pattern of the Cholesky matrix $\mCholPrec$; only colored pixels have non-zero values. Elements in blue correspond to (positive) diagonal terms, while the elements in pink, correspond to off-diagonal values.
    (Right)~The corresponding sparsity pattern of the precision matrix $\mPrecision$, where $\mPrecision= \mCholPrec \trans{\mCholPrec}$.}
    \label{fig:sparsity_pattern}
\end{figure*}

\section{Training details}
We use the Adam \cite{kingma2015adam} optimiser with an initial learning rate of $1e^{-4}$, which we halve after the first, fifth and 15th epoch. We train for 20 epochs in total. A batch size of 16 images was used for all experiments. Our initial encoder and (mean) decoder are pre-trained models from the ensemble, we do not use samples from that model as observations during the training process.

\section{Pixelwise summaries}
\begin{figure*}[h!]
    \centering
    \renewcommand{\arraystretch}{0.3}
    \begin{adjustbox}{max width=\textwidth}
    \begin{tabular}{cc}
        \includegraphics[width=0.5\textwidth]{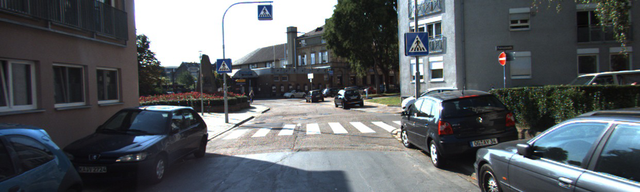} & \\
        Boot+Log & SUPN Boot+Log \\
        \includegraphics[width=0.5\textwidth]{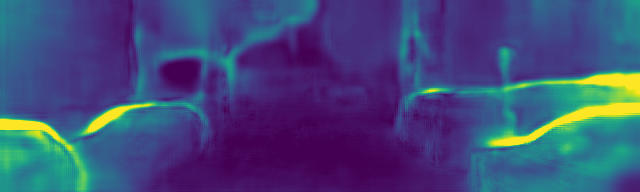}
        &
        \includegraphics[width=0.5\textwidth]{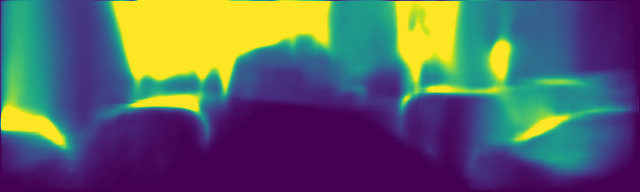}
        \\
        \includegraphics[width=0.5\textwidth]{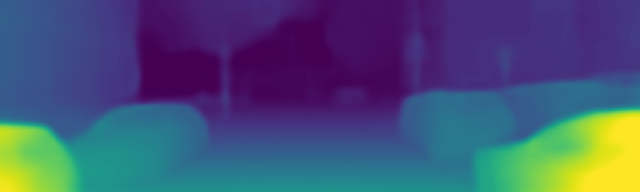}
        &
        \includegraphics[width=0.5\textwidth]{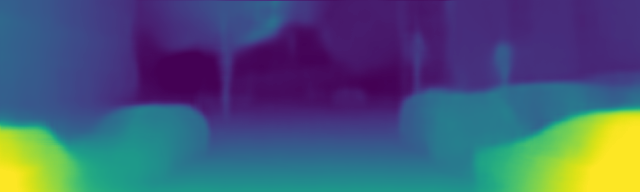}
        \\
        \includegraphics[width=0.5\textwidth]{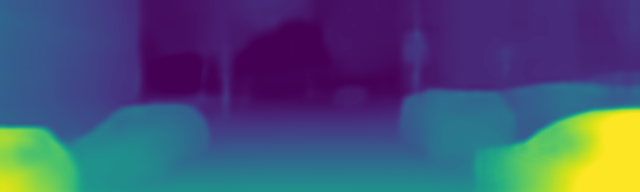}
        &
        \includegraphics[width=0.5\textwidth]{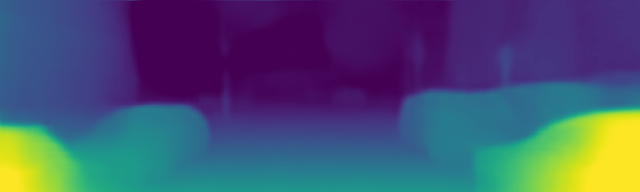}
        \\
        \includegraphics[width=0.5\textwidth]{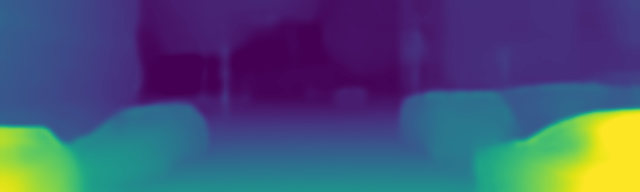}
        &
        \includegraphics[width=0.5\textwidth]{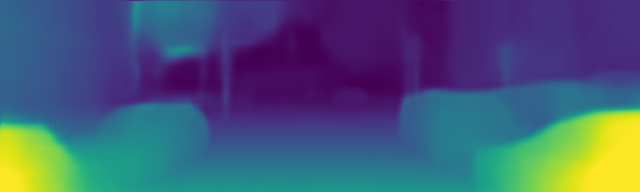}
        \\
        \includegraphics[width=0.5\textwidth]{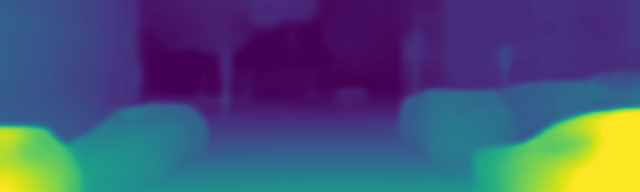}
        &
        \includegraphics[width=0.5\textwidth]{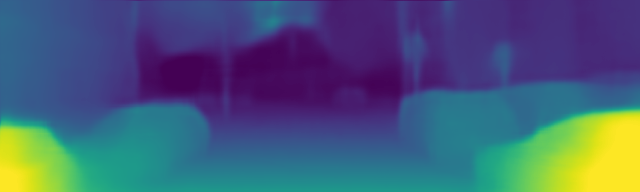}
        \\
        \includegraphics[width=0.5\textwidth]{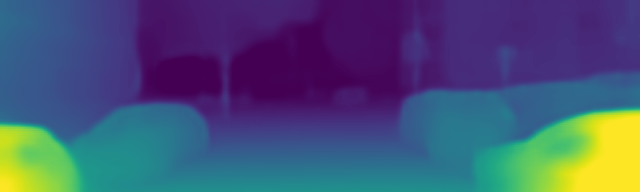}
        &
        \includegraphics[width=0.5\textwidth]{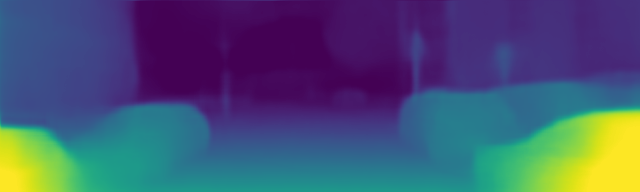}
    \end{tabular}
    \end{adjustbox}
    \caption{Visualisation of standard deviations (2nd row) and disparity samples (3rd row and below) for a given image (top row). Both of the standard deviations, and all of the samples are normalised to be in the same range. 
    All of the samples are spatially smooth and we note that our predicted standard deviations show similar structure, although the map is smoother than that derived from the ensemble. We also observe substantially more uncertainty about the sky pixels with our approach. We speculate this is due to the inherent uncertainty in the monocular depth prediction task, which induces large variability between the ensemble models across images that the SUPN model has tried to faithfully capture.}
\end{figure*}

\newpage

\section{Dataset and code}

\begin{description}
    \item[KITTI dataset \cite{geiger2013vision}] Available from \url{http://www.cvlibs.net/datasets/kitti/}. Available for non-commercial use only. License: \href{https://creativecommons.org/licenses/by-nc-sa/3.0/}{Creative Commons Attribution-NonCommercial-ShareAlike 3.0}
    
    \item[Monodepth2 repository \cite{godard2017monodepth}] Available from \url{https://github.com/nianticlabs/monodepth2}. Available for non-commercial use only. License: \url{https://github.com/nianticlabs/monodepth2/blob/master/LICENSE}
    
    \item [Poggi et al repository \cite{poggi2020uncertainty}] Available from: \url{https://github.com/mattpoggi/mono-uncertainty}. License: \href{https://github.com/mattpoggi/mono-uncertainty/blob/master/evaluate.py}{MIT License}
    
    \item[Pytorch] Available from: \url{https://pytorch.org/} License: \url{https://github.com/pytorch/pytorch/blob/master/LICENSE}
    
    \item [SuiteSparse] Available from: \url{https://github.com/DrTimothyAldenDavis/SuiteSparse}. License: LGPL-2.1

    \item [torch-sparse-solve 0.0.5] Available from: \url{https://pypi.org/project/torch-sparse-solve/} License: LGPL-2.1
\end{description}

\end{document}